%% file: main.tex
\documentclass[10pt,twocolumn,letterpaper]{article}

\usepackage[pagenumbers]{cvpr} 

\usepackage{multirow}
\usepackage{soul}
\usepackage{placeins}

\input{preamble}
\definecolor{cvprblue}{rgb}{0.21,0.49,0.74}
\usepackage[pagebackref,breaklinks,colorlinks,allcolors=cvprblue]{hyperref}

\def\paperID{}
\def\confName{CVPR}
\def\confYear{2026}

\title{FedVG: Gradient-Guided Aggregation for Enhanced Federated Learning}

\author{
Alina Devkota$^{1}$ \quad
Jacob Thrasher$^{1}$ \quad
Donald Adjeroh$^{1}$ \quad
Binod Bhattarai$^{2, 3}$ \quad
Prashnna K.~Gyawali$^{1}$\\[0.4em]
West Virginia University$^{1}$,
University of Aberdeen$^{2}$,
Fogsphere (Redev.AI)$^{3}$\\[0.4em]
{\tt\small
\{ad00139,jdt0025\}@mix.wvu.edu,
\{donald.adjeroh,prashnna.gyawali\}@mail.wvu.edu,}\\
{\tt\small
binod.bhattarai@abdn.ac.uk}
}

\begin{document}
\maketitle

\vspace{-15pt}
\input{sec/0_abstract}    
\vspace{-4pt}


\input{sec/1_intro}
\vspace{-4pt}
\input{sec/2_related}

\vspace{-4pt}
\input{sec/3_method}

\vspace{-4pt}
\input{sec/4_experiments}
\vspace{-4pt}
\input{sec/5_results}
\vspace{-4pt}
\input{sec/6_conclusion}
\vspace{-4pt}
\newpage
{
    \small
    \bibliographystyle{ieeenat_fullname}
    \bibliography{main}
}

\input{sec/X_suppl}

\end{document}

%% file: sec/0_abstract.tex
\begin{abstract}
Federated Learning (FL) enables collaborative model training across multiple clients without sharing their private data. 
However, data heterogeneity across clients leads to client drift, which degrades the overall generalization performance of the model.
This effect is further compounded by overemphasis on poorly performing clients.
To address this problem, we propose FedVG, a novel gradient-based federated aggregation framework that leverages a global validation set to guide the optimization process. Such a global validation set can be established using readily available public datasets, ensuring accessibility and consistency across clients without compromising privacy.
In contrast to conventional approaches that prioritize client dataset volume, FedVG assesses the generalization ability of client models by measuring the magnitude of validation gradients across layers. 
Specifically, we compute layerwise gradient norms to derive a client-specific score that reflects how much each client needs to adjust for improved generalization on the 
global
validation set, thereby enabling more informed and adaptive federated aggregation. Extensive experiments on both natural and medical image benchmarking datasets, across diverse model architectures, demonstrate that FedVG consistently improves performance, particularly in highly heterogeneous settings.
Moreover, FedVG is modular and can be seamlessly integrated with various state-of-the-art FL algorithms, often further improving their results.
Project Page: \href{https://machine-intelligence-lab-wvu.github.io/fedvg/}{https://machine-intelligence-lab-wvu.github.io/fedvg/}.
\end{abstract}

%% file: sec/1_intro.tex
\section{Introduction}
\label{sec:intro}
\vspace{-5pt}

\begin{figure}[t]
\centering
\includegraphics[width=\columnwidth]{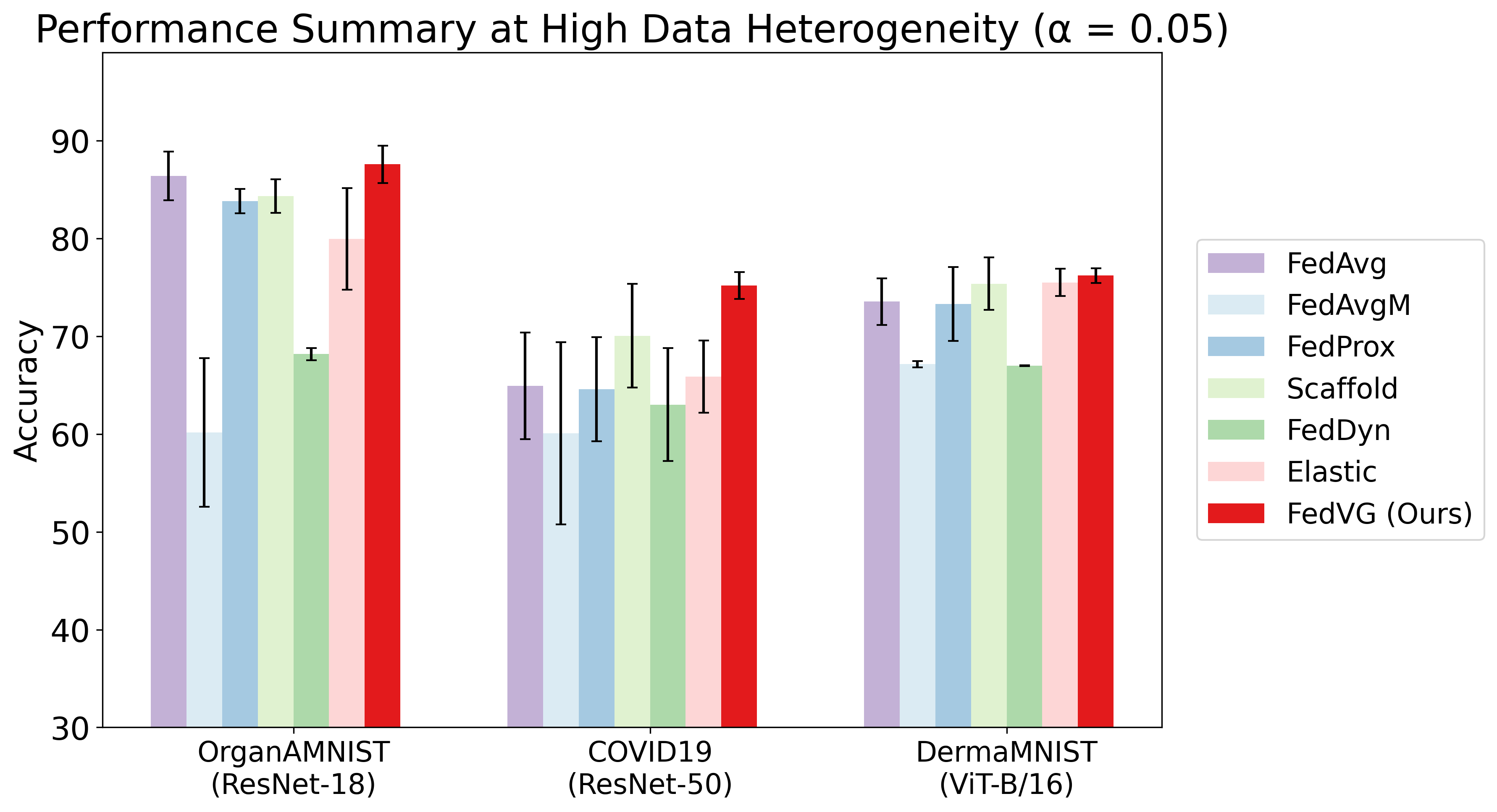}
\caption{Comparison of federated learning methods under high data heterogeneity ($\alpha = 0.05$). FedVG consistently achieves the highest or near-highest accuracy across all settings.}
\label{fig:intro}
\end{figure}

Federated Learning (FL) has emerged as a powerful privacy-preserving paradigm for collaboratively training machine learning models across decentralized clients. Its relevance is rapidly growing in domains like healthcare, where sensitive patient data must remain localized, yet robust models are needed for diagnostics, prognosis, or medical imaging \cite{thrasher2023multimodal, xu2021federated}.

In an FL setup, a central server coordinates the training of a shared global model by aggregating model updates from multiple clients. Each client locally trains a copy of the global model on its own dataset.
After local training, the server performs an aggregation step, which is commonly an average of the client parameters, weighted by their dataset size.
These methods assume that weighting by dataset size 
sufficiently reflects each client’s contribution.

However, such a naive reliance on clients' data volume overlooks deeper behavior
during training and fails to address client heterogeneity, a fundamental challenge in FL \cite{wang2020tacklingobjectiveinconsistencyproblem}. 
Clients often differ significantly in terms of data distribution, model initialization, and computational resources.
This non-IID nature exacerbates client drift \cite{karimireddy2020scaffold}, where local models diverge,
degrading convergence and resulting in poor global generalization. Despite FL’s privacy-preserving appeal, such heterogeneity limits its practical deployment. 

To address this, we propose \textbf{Fed}erated aggregation via \textbf{V}alidation \textbf{G}radients (\textbf{FedVG}), a more informed aggregation strategy that avoids the naive assumption of client uniformity and instead leverages the training dynamics of each client. Specifically, we focus on gradient analysis to evaluate clients during local training and how beneficial their updates are to the global model. Since gradients reflect the direction and magnitude of optimization progress, they serve as a rich signal for understanding training behavior.

However, local gradients alone may be biased by the client’s own data distribution.
To better assess generalization and mitigate this bias, we introduce a client-agnostic validation set that remains fixed across training rounds.
This validation set can be constructed from readily available public datasets that share similar characteristics with the target domain, such as similar imaging modalities, classes, etc.
Leveraging such public dataset, which is not biased toward any single client, can serve as common and neutral reference point \cite{koch2021reduced} to evaluate model updates
without requiring access to any private data.
By computing gradients of each client’s model with respect to this shared validation set, we obtain a more objective signal that reflects how well the model performs on data outside its local distribution.


Prior work has shown that models situated in optimal or sub-optimal regions of the loss landscape tend to exhibit flatter gradients, characterized by smaller gradient norms
\cite{hochreiter1994simplifying, chaudhari2019entropy, ly2017tutorialfisherinformation}. In contrast, models in sharper regions, typically overfitted or less confident, produce larger gradients, indicating a need for significant updates to reach more generalizable solutions.
Gradient magnitudes can also serve as a proxy for epistemic uncertainty in model predictions \cite{wang2024epistemic}. Thus, by measuring the norm of the validation gradients, we can score clients based on the stability and confidence of their models, and weigh their updates accordingly. This enables a more adaptive and principled aggregation process that emphasizes generalization over local performance.


We conduct a comprehensive evaluation of FedVG across multiple datasets using various backbone models and levels of data heterogeneity. To further assess its effectiveness, we analyze performance under distribution shifts in the global validation set, explore different aggregation granularities, and evaluate the impact of the proposed weighting scheme. Results demonstrate FedVG's robustness across a wide range of settings, particularly under high heterogeneity (as shown in Fig. \ref{fig:intro}). Additionally, we show that FedVG is a modular, pluggable approach that integrates seamlessly with existing FL algorithms to enhance their performance.

Overall, our contributions are as
follows: (1) We propose \textbf{FedVG}, a novel gradient-based aggregation method that leverages a global validation set to assign higher weights to clients with flatter validation gradients, promoting better generalization. (2) We extensively evaluate FedVG on multiple datasets, diverse model architectures, and varying levels of data heterogeneity. Additional analysis includes distribution shifts, aggregation granularity, and the impact of the weighting scheme. (3) We show that FedVG is a pluggable module that can be seamlessly integrated with existing FL algorithms to enhance their performance.

%% file: sec/2_related.tex
\section{Related Works}
\label{sec:related}
\vspace{-5pt}
The vanilla FedAvg algorithm aggregates client updates weighted by local dataset sizes \cite{mcmahan2017communication}, but suffers under non-iid settings. To address this, methods like FedProx \cite{li2020federated}, and Scaffold \cite{karimireddy2020scaffold} add regularization to reduce divergence, while
FedAvgM \cite{hsu2019measuring} adds server-side momentum.
Gradient-based aggregation offers directional signals beyond scalar metrics. FedDyn~\cite{acar2021federated} dynamically aligns local and global stationary points of the loss function.
Other works use temporal gradient similarity with global model \cite{zang2024efficient} or mutate global model \cite{hu2024fedmut} for adaptive weighing.
Unlike these approaches, which focus on model gradients or similarity heuristics, FedVG uses validation gradients to assess generalization quality of each client update.

Some works leverage shared or auxiliary data to reduce distribution skew. FCCL \cite{zhang2023target} leverages unlabeled or synthetic data to align class representations across clients, while \cite{zhao2018federated} shares a small IID dataset to mitigate heterogeneity. Conversely, other works utilize public data for federated transfer learning. For example, FedMD \cite{li2019fedmdheterogenousfederatedlearning} utilizes a large public dataset to enhance the performance of clients with limited local data. Cronus \cite{chang2019cronusrobustheterogeneouscollaborative} and FedH2L \cite{li2021fedh2lfederatedlearningmodel} instead utilize auxiliary data to enhance security by essentially obfuscating client parameters during aggregation. Finally, FedDF \cite{lin2021ensembledistillationrobustmodel} applies knowledge distillation to support federated environments with heterogeneous client architectures.
FedVG also uses a global shared dataset, but instead of using it for training, it is used for computing validation gradients.
Unlike these works, FedVG works effectively with non-IID validation sets and adapts weights dynamically.

Research shows that model layers diverge differently across clients. Both shallow and deep layers exhibit significant variation, especially under noisy or low-quality data \cite{zhao2018federated, tam2023federated, chen2023layer, lee2024layer}.
Deeper layers tend to diverge more strongly,
causing misalignment in representations and reduced model robustness. 
FedNCL \cite{tam2023federated} addresses this by detecting unreliable clients and performing layerwise aggregation.
Similarly, FedMA \cite{wang2020federated}, and  FedLAMA \cite{lee2023layer} independently aggregate each layer to improve global alignment.

Validation gradients have been applied for tasks such as hyperparameter tuning \cite{barratt2018optimizing}, adapting regularization \cite{brito2025cross}, or architecture search \cite{li2020neural}.
However, \cite{chen2023layer} shows that gradients vary significantly across layers: shallow layers tend to converge to flatter minima with more stable gradients, while deeper layers are more sensitive.
Motivated by these findings, FedVG incorporates gradient information from all layers, allowing for more holistic adaptation to better capture client model alignment and robustness. 
This aligns with insights from sharpness-aware training methods
like SAM \cite{foret2020sharpness}, which favors flatter minima to improve robustness.

To summarize, while prior works have explored layerwise assessment or gradient alignment separately, the use of validation gradient to guide client weighting remains underexplored in FL. FedVG fills this gap by leveraging a global validation set to compute per-layer gradient norms, dynamically assigning higher weights to clients whose updates lie in flatter, more generalizable regions of the loss surface.

%% file: sec/3_method.tex
\begin{figure*}[t]
\centering
\includegraphics[width=.9\textwidth]
{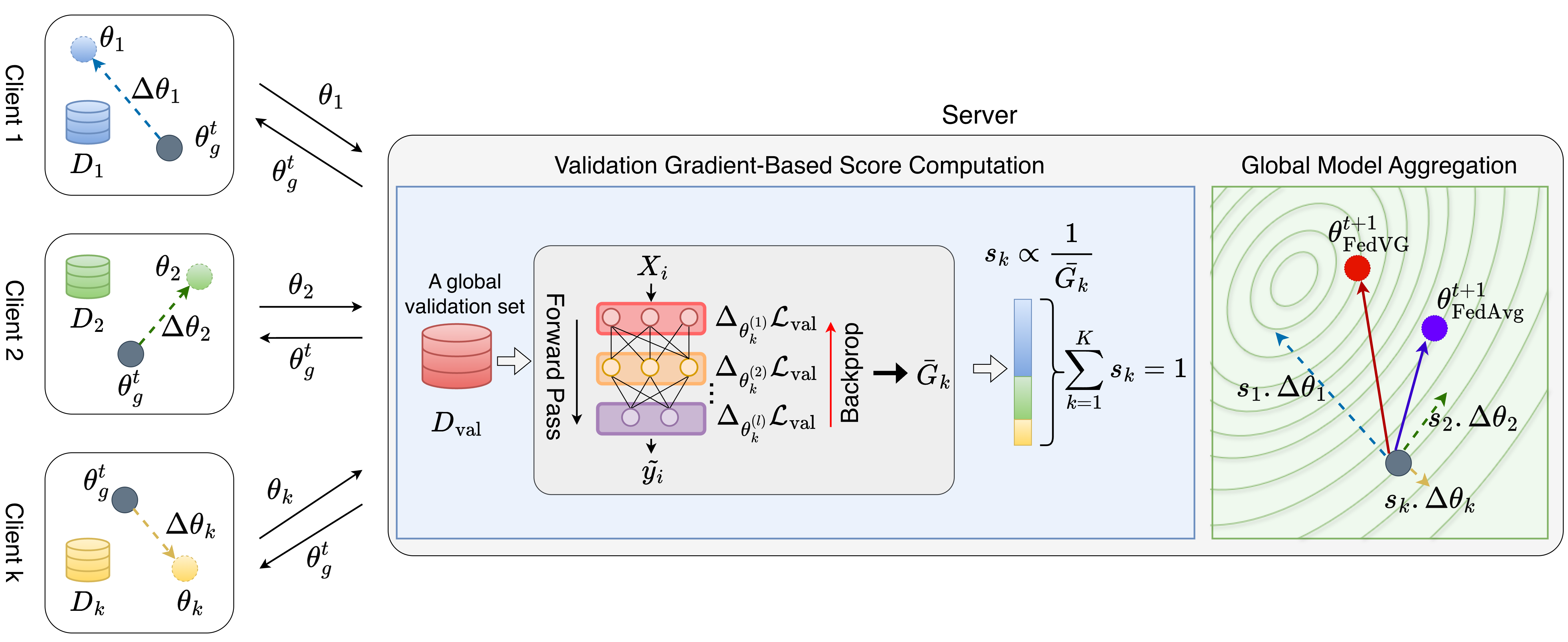}
\caption{\textbf{FedVG Framework}. Locally trained client models, which converge to varying points, are sent to the server. A global validation set is used to compute their validation gradients, obtain client scores $s_k$, and form a global model by weighted sum. Individual client updates are represented as dotted lines in the loss landscape on the right, while aggregated models are solid.
}
\label{fig:methodology}
\end{figure*}

\section{Methodology}
\vspace{-5pt}

\subsection{Problem Formulation}
\vspace{-4pt}
We consider a standard federated learning setup with $K$ clients. Each client $k \in \{1,2,3 ... K\}$ holds a private local training dataset $D_k$ of size $n_k$ and maintains a local model $\theta_k$.
The overall training dataset $D$ is the union of all client datasets, i.e. $D = \cup_{k=1}^K{D_k}$ and total number of training samples is $N = \sum_{k=1}^{K}n_k$.
During each communication round, each client $k$ trains its local model $\theta_k$ using its local dataset $D_k$ and sends
the updates 
$\Delta \theta_1, \Delta \theta_2, ... \Delta \theta_K$
to the central server. The server aggregates these model updates to form a global model $\theta_g$, which is then
distributed back to the clients for the next round of training.

Various strategies can be used to aggregate client updates. 
A common approach is
FedAvg \cite{mcmahan2017communication}, which weighs each client's update ($\Delta\theta_k = \theta_g - \theta_k$) by its data proportion, i.e., $\frac{n_k}{N}$, during aggregation.

\begin{align}
    \theta_g \leftarrow \theta_g - \sum_{k=1}^{K} \frac{n_k}{N} \Delta\theta_k
\end{align}

Under homogeneous client distributions, weighing client updates by dataset size is reasonable, as larger datasets typically yield more reliable local models. However, in heterogeneous settings, local model quality may not correlate with dataset size. 
This limitation has motivated the development of alternative aggregation strategies that assign client weights based on additional factors beyond data volume.

Following \cite{li2020federated}, we define a general client-specific score $s_k \in \left[ 0, 1 \right]$, which quantifies the weight assigned to a client $k$ by a given federated algorithm.

\begin{equation}
\theta_g \leftarrow \theta_g - \sum_{k=1}^{K} s_k \Delta \theta_k
\label{eq:fl-agg}
\end{equation}

where, $\sum_{k=1}^K{s_k} = 1$. This formulation allows for flexible integration of various strategies, including data-size-based weighting (as in FedAvg \cite{mcmahan2017communication}), client loss-based weighting \cite{wang2020tackling}, or gradient alignment methods (Scaffold \cite{karimireddy2020scaffold}, FedAC \cite{zang2024efficient}). 
In our approach, we use gradient norms computed on a shared global validation dataset to compute the aggregation weights $s_k$.

\vspace{-4pt}
\subsection{Validation Gradient-based aggregation}
\vspace{-4pt}
FedVG is a gradient-based FL aggregation framework that leverages a publicly available global validation set to guide the aggregation of client updates.
At the server, this dataset is used to evaluate each client’s generalization ability.
While validation loss provides a coarse measure of performance, prior work has shown that the classifier head often absorbs most of the local dataset bias \cite{luo2021no}, causing validation loss to \textit{overemphasize} the performance of this final layer while overlooking discrepancies in deeper representations that influence generalization.
Validation gradients, in contrast, provide richer information by capturing not only how well a model performs but also how its parameters should change to improve generalization.
FedVG thus adaptively weighs client updates according to the magnitude of their validation gradients, prioritizing those that contribute most to generalization and mitigating the challenges posed by data heterogeneity across clients.

An overview of FedVG is shown in Fig.~\ref{fig:methodology}.
In this example, client models converge to varying points (left), as demonstrated by their individual update vectors.
FedVG computes validation gradient-based scores for each client and scales their contributions accordingly, visualized in the loss landscape (right).
Traditional methods such as FedAvg perform a simple average of updates, often placing the global model in suboptimal, high-loss regions.
As shown in the figure, the FedAvg solution drifts away from the low-loss valley, whereas the FedVG solution, guided by validation gradient information, lies closer to the region of minimal validation loss.
This highlights how validation gradients can effectively capture clients’ generalization quality and provide a principled basis for adaptive aggregation.

Below, we describe our gradient estimation strategy and how it is used for client score computation, establish its theoretical connection to the Fisher Information Matrix, and discuss the modular design of FedVG.

\vspace{-4pt}
\subsubsection{Gradient Estimation and Score Computation}
\vspace{-4pt}
Steeper gradients indicate sharper regions in the loss landscape, whereas smaller gradients correspond to flatter regions that are typically associated with better generalization \cite{foret2020sharpness}.
This observation motivates the core idea behind FedVG: clients whose models exhibit flatter validation landscapes (i.e., smaller validation gradients) are likely to generalize better and should therefore be assigned higher influence during aggregation.
Moreover, layers exhibit distinct behaviors, especially under non-IID settings \cite{chen2023layer, lee2024layer}, meaning that different parts of the model may generalize differently across clients.
To account for this, FedVG aggregates gradient information across all layers to obtain a holistic and structurally aware measure of each client’s generalization alignment.

Formally, we introduce a shared validation dataset, $D_{\mathrm{val}} = \{(x_i, y_i)\}_{i=1}^{N_{\mathrm{val}}},$
where $x_i \in \mathcal{X}$ and $y_i \in \mathcal{Y}$ follow the same data structure and task as the client training datasets $D$.
After each client $k$ completes local training on $D_k$,
we evaluate the trained client model (with parameters $\theta_k$) on the global validation dataset $D_{\mathrm{val}}$. 
For a deep neural network, the parameter vector can be decomposed into $L$ layers, $\theta_k = \{\theta_k^{(1)}, \theta_k^{(2)}, \ldots, \theta_k^{(L)}\}$,
which allows us to compute layerwise validation gradients and assess each layer's contribution to the global alignment.
For each client $k$, we define the validation loss as
\begin{equation}
    \mathcal{L}_{\mathrm{val}}(\theta_k) = \frac{1}{N_{\mathrm{val}}} 
    \sum_{(x_i, y_i) \in D_{\mathrm{val}}} \ell \big( f(x_i; \theta_k), y_i \big),
\end{equation}
where $\ell(\cdot)$ is the task-specific loss function (e.g., cross-entropy), and $f(x; \theta_k)$ denotes the client model’s output.

We then compute the validation loss gradients with respect to each layer $\ell$ of the client model parameters:
$
\nabla_{\theta_k^{(\ell)}} \mathcal{L}_{\mathrm{val}}
$.
This is achieved via a standard backward pass.
As discussed earlier, layerwise gradients reveal each layer's sensitivity to global validation data, exposing misalignments between client representations and the global distribution, particularly in deeper layers, more affected by local data bias (also demonstrated in the Appendix).

Finally, we aggregate the layerwise gradients into a single scalar measure for each client by 
averaging the norms of the gradients from each layer:
\begin{equation}
    \bar{G}_k = \frac{1}{L}\sum_{\ell=1}^{L}{\left\|  \nabla_{\theta_k^{(\ell)}} \mathcal{L}_{\mathrm{val}} \right\|}.
    \label{eq:general_norm}
\end{equation}
where $\left\| \cdot \right\|$ denotes a general vector norm.

After computing the average gradient norm $\bar{G}_k$ for each client $k$, the next step is to derive a corresponding score that determines its contribution to global model aggregation. 
A smaller validation gradient norm indicates that the client's model parameters are already well aligned with the global distribution, requiring minimal adjustment to further reduce the validation loss. Consequently, clients with lower gradient norms inherently exhibit stronger generalization and are assigned higher aggregation scores, defined as:

\begin{equation}
s_k = \frac{1 / (\bar{G}_k + \epsilon)}{\sum_{j=1}^{K} 1 / (\bar{G}_j + \epsilon)}
\label{eq:score_fedvg}
\end{equation}
where $\epsilon$ is a small constant added for numerical stability.
These scores $s_k$ are then used in Eqn. \ref{eq:fl-agg} to weigh the model updates from each client during server-side aggregation. 
\vspace{-3pt}
\subsubsection{Relationship to Fisher Information}
\vspace{-3pt}
FedVG's weighting mechanism can be formally understood through the lens of the Fisher Information Matrix (FIM) \cite{Fisher1922-ei}, which quantifies the local curvature of the loss surface with respect to model parameters. 
Establishing this connection is important because it provides a theoretical foundation for using validation gradients as indicators of generalization: clients with smaller Fisher information (i.e., flatter curvature) correspond to models that lie in broader minima and thus generalize better.
This interpretation aligns closely with FedVG’s design, where smaller validation gradient norms are assigned higher aggregation weights.


The gradient of the cross-entropy loss is given by
\begin{equation}
\nabla_\theta \mathcal{L}_{\mathrm{CE}}(\theta) = - \nabla_\theta \log p_\theta(y|x),
\end{equation}
which corresponds to the negative score function, whose second-order moment is the FIM:
\begin{equation}
F(\theta) = \mathbb{E}_{(x,y) \sim D} \Big[ \nabla_\theta \log p_\theta(y|x) \, \nabla_\theta \log p_\theta(y|x)^\top \Big].
\end{equation}
The FIM measures how sensitive the model’s likelihood is to small parameter changes. Large Fisher Information indicates a sharp region, where small parameter perturbations cause large changes in the output distribution, while small Fisher Information corresponds to a flat, stable region. \cite{ly2017tutorialfisherinformation}

FedVG quantifies each client’s sensitivity using the mean layerwise norm of its validation gradients (Eq. ~\ref{eq:general_norm}). Since the gradient of the cross-entropy loss directly corresponds to the score function, $\|\nabla_\theta \mathcal{L}_{\mathrm{CE}}\|_2$ can be interpreted as the square root of Joint Fisher \cite{pmlr-v267-li25j}, a diagonal approximation to the Fisher Information, 
and $\|\nabla_\theta \mathcal{L}_{\mathrm{CE}}\|_1$ is directly proportional to square root of the Joint Fisher.
Both norms measure the local sensitivity of the model parameters. 


\vspace{-4pt}
\subsubsection{Modular Integration with existing FL Algorithms}
\vspace{-4pt}
Another key aspect of FedVG is its modularity. The gradient-based scoring 
integrates seamlessly into existing FL frameworks,
enabling broad applicability without major changes to the FL pipeline.
We first present the general aggregation rule used in most FL algorithms as:
\begin{equation}
\theta_{\text{g}}^{t+1} = \left( \theta_g^t - \sum_{k=1}^{K} s_k \cdot \Delta\theta_k^{t+1} \right) + \mathcal{R}^{t+1}
\label{eq:fl-general}
\end{equation}
where, $\theta_g^t$ is the global model at round $t$, $\Delta \theta_k^{t+1} = \theta_k^{t+1} - \theta_g^t$ is client $k$'s local update, $s_k$ its weight, and $\mathcal{R}^{t+1}$ is the regularization term specific to the FL method. 
Many algorithms like Fedprox, FedDyn
use $s_k = \frac{n_k}{N}$ similar to FedAvg. 
For regularization, FedAvg sets $\mathcal{R}^{t+1} = 0$, FedDyn uses $\mathcal{R}^{t+1} = \beta(\theta_g^{t+1} - \theta_g^t)$, and methods like FedAvgM and Scaffold include momentum, or control variate corrections within $\mathcal{R}^{t+1}$.
FedVG fits into this framework by replacing the client weights $s_k$ with our gradient-driven scores from Eqn.  \ref{eq:score_fedvg}, or by averaging these scores with existing weights.


\begin{figure*}[t]
\centering
\includegraphics[width=\textwidth]{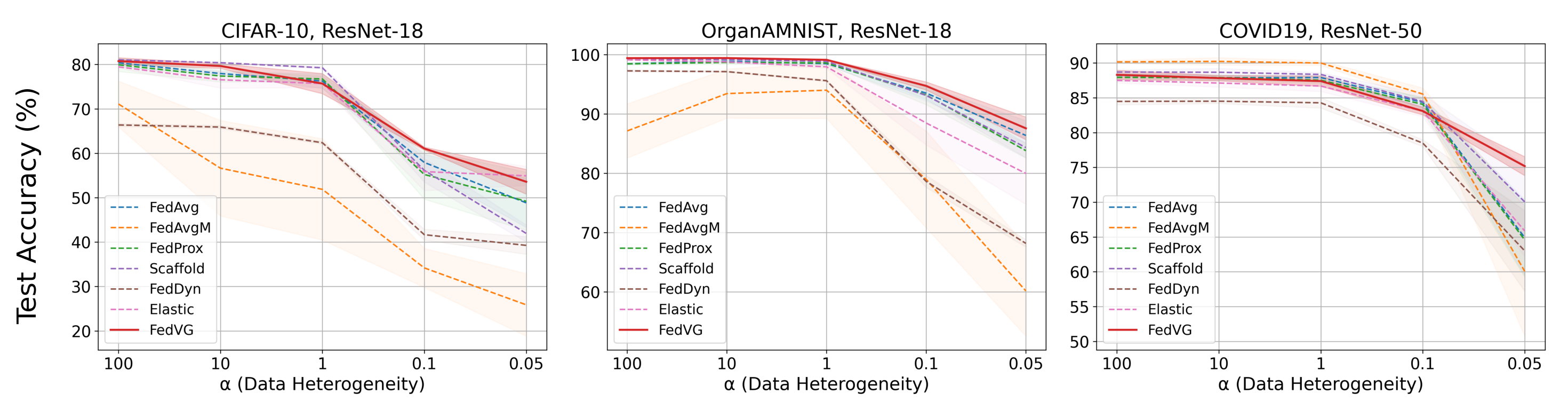}
\caption{FL algorithm performance as $\alpha \rightarrow{} 0$. Shaded areas indicate standard deviations. Additional results for TinyImageNet and DermaMNIST datasets on the ResNet-50 model are provided in the Appendix.}
\label{fig:het-levels}
\end{figure*}

%% file: sec/4_experiments.tex
\section{Experiments}
\vspace{-5pt}
We conduct extensive experiments to evaluate the effectiveness of FedVG across both natural and medical image classification tasks and on different model architectures under varying levels of heterogeneity. In this section, we provide a detailed overview of our experimental setup.

\vspace{-4pt}
\subsection{Datasets}
\vspace{-4pt}
We evaluate the performance of FedVG using the benchmark classification datasets: \textbf{CIFAR-10} 
and \textbf{TinyImageNet}. 
To further validate its effectiveness, we also analyze three additional medical imaging datasets: \textbf{OrganAMNIST} 
\cite{medmnistv2}, \textbf{COVID19} 
\cite{sait2020curated}, and \textbf{DermaMNIST} 
\cite{medmnistv2}. 
OrganAMNIST consists of 58,830 axial-view abdominal CT scans labeled for multi-class classification across 11 body organs. 
The COVID19 dataset includes 9,523 chest X-rays labeled as healthy, COVID-19, viral pneumonia, or bacterial pneumonia. 
Finally, DermaMNIST is 
a dermatology image dataset
comprising 10,015 images spanning 7 diagnostic classes.
Overall, these datasets
encompass a diverse set of classification tasks, 
varying in image resolution, class number, and domain complexity.

\vspace{-4pt}
\subsection{Experimental Setting}
\vspace{-4pt}
We use ResNet-18 \cite{he2015deepresiduallearningimage} as the backbone model for experiments on CIFAR-10 and OrganAMNIST, and ResNet-50 for the higher-resolution datasets: TinyImageNet, COVID19, and DermaMNIST.
We also analyzed transformer architectures using the ViT backbone \cite{dosovitskiy2020image}, employing ViT-S/16 for COVID19 and ViT-B/16 for DermaMNIST.
To maintain consistency with existing FL benchmarks, we implemented our federated setup using the FL-Bench framework \cite{Tan_FL-bench}.
For each experiment, we allocate 10\% of the dataset for $D_{val}$ and 25\% as the test set. The remaining data is distributed across the $K$ clients using a Dirichlet distribution \cite{hsu2019measuring} parameterized by $\alpha$,
where a smaller $\alpha$ results in a highly heterogeneous distribution, and a larger $\alpha$ produces a more homogeneous distribution.
To evaluate the performance of FedVG under varying degrees of data heterogeneity, we conduct experiments with $\alpha \in \{100, 10, 1, 0.1, 0.05\}$.
We conduct 200 rounds of federated training,
randomly sampling $p\%$ of clients at each round for training.
Each selected client trains its local model for 5 epochs using the SGD optimizer. 
All experiments were performed on NVIDIA A30 GPUs with results averaged over 5 random seeds.
Additional configuration details are provided in the Appendix.


\vspace{-4pt}
\subsection{Baselines}
\vspace{-4pt}
We compare FedVG against several established federated learning algorithms. 
\textbf{FedAvg} \cite{mcmahan2017communication} averages client models without addressing client drift. 
\textbf{FedAvgM} \cite{hsu2019measuring} adds server-side momentum to FedAvg.
\textbf{FedProx} \cite{li2020federated} adds a proximal term in the local objective to penalize large deviations from the global model. 
\textbf{Scaffold} \cite{karimireddy2020scaffold} uses control variates to correct client drift.
\textbf{FedDyn} \cite{acar2021federated} applies dynamic regularization to align client optima with global stationary points.
\textbf{Elastic} aggregation \cite{chen2023elastic} adaptively interpolates client 
updates based on parameter sensitivity. 
All baselines are implemented in the same framework and evaluated under identical data splits, model architectures, and training protocols.

%% file: sec/5_results.tex
\section{Results}
\vspace{-5pt}
\label{sec:results}
\subsection{Performance at varying heterogeneity}
\vspace{-4pt}


We present our primary result in Fig. \ref{fig:het-levels}. Due to space restrictions, we include results for three datasets (CIFAR-10, OrganAMNIST, and COVID-19) evaluated against all baselines. Results on Tiny-ImageNet and DermaMNIST can be found in the Appendix.
As demonstrated, FedVG consistently 
achieves competitive or superior performance,
particularly under high data heterogeneity (i.e., low $\alpha$).
On CIFAR-10,
FedVG maintains strong accuracy across all $\alpha$ levels, excelling at $\alpha = 0.05$ and outperforming FedAvg by a notable margin. 
For OrganAMNIST,
FedVG consistently outperforms all baselines across all values of $\alpha$.
On COVID19,
FedVG performs competitively overall and notably surpasses all baselines at $\alpha = 0.05$, highlighting its robustness under high heterogeneity.
In terms of variability, FedVG consistently maintains low standard deviation across all datasets, indicating stable performance. 

\begin{figure}[t]
\centering
\includegraphics[width=\columnwidth]{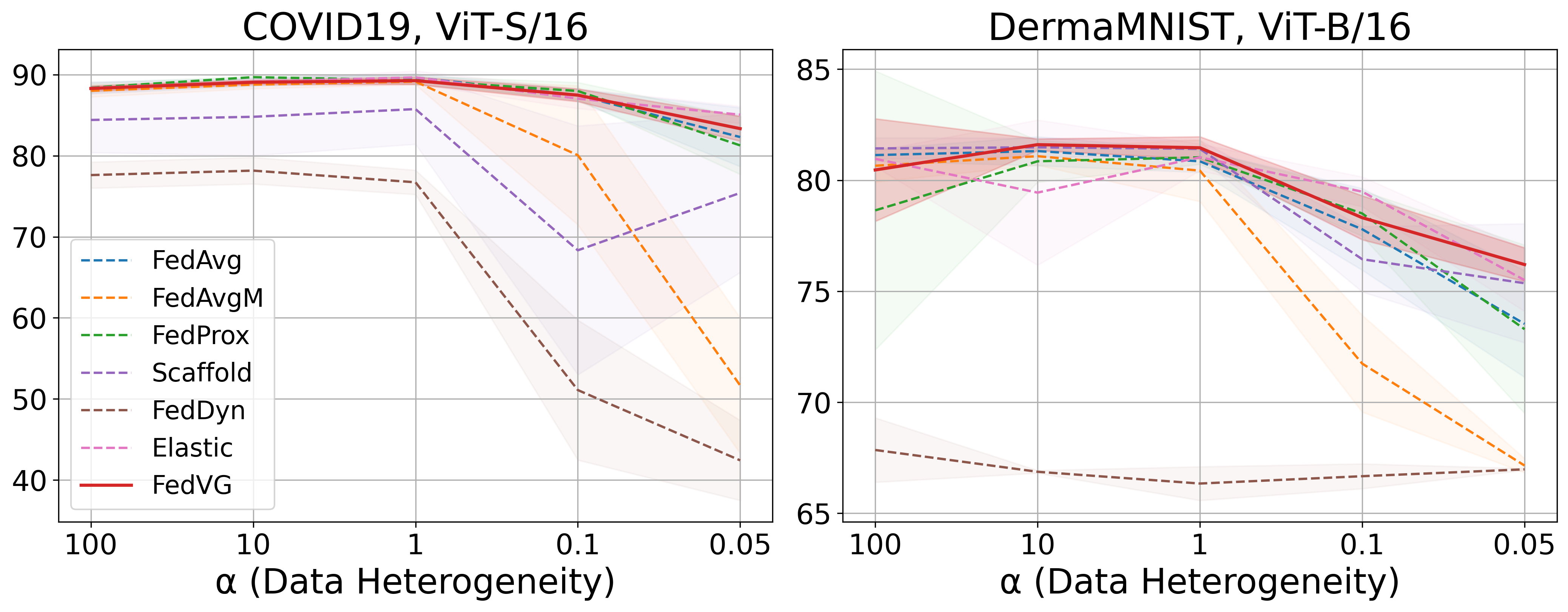}
\caption{FL algorithm performance on ViT models as $\alpha \rightarrow{} 0$. Shaded areas show standard deviations.
}
\label{fig:results_vit}
\end{figure}

To evaluate the adaptability of FedVG to modern architectures, we extend our experiments to Vision Transformers, specifically using ViT-S/16 and ViT-B/16 models on the higher-resolution datasets in our study-namely, COVID19 and DermaMNIST. 
As shown in Fig.~\ref{fig:results_vit}, FedVG consistently ranks among the top-performing methods on transformer backbones across both datasets, particularly under high heterogeneity.
These results highlight FedVG's scalability beyond CNNs and its effectiveness in guiding model updates in federated settings involving high-capacity, non-convolutional architectures. 

Furthermore, we conducted statistical significance testing using the  Wilcoxon signed-rank test to evaluate the performance differences between FedVG and the baselines. 
At a significance level of $p<0.05$, 
FedVG consistently outperforms FedDyn across all heterogeneity levels 
In addition, 
FedVG shows statistically significant improvements over 
Elastic at $\alpha \in \{ 10, 1\}$, 
Scaffold at $\alpha \in \{ 0.1, 0.05 \}$, FedAvgM at $\alpha \in \{ 0.1, 0.05 \}$, and both FedAvg and FedProx at $\alpha \in \{0.05 \}$.
Importantly, none of the baselines significantly outperform FedVG at any $\alpha$ value.
While not statistically significant according to the Wilcoxon test, FedVG still achieves higher average performance than all baselines under extreme heterogeneity ($\alpha = 0.05$).

Overall, our results indicate that FedVG generalizes well across datasets and architectures, demonstrating wide applicability. On medical datasets, it consistently ranks among top performers, indicating robustness to non-IID client data and potential for real-world federated deployments. Further details and additional results are provided in the Appendix.

\begin{figure}[t]
\centering
\includegraphics[width=\columnwidth]{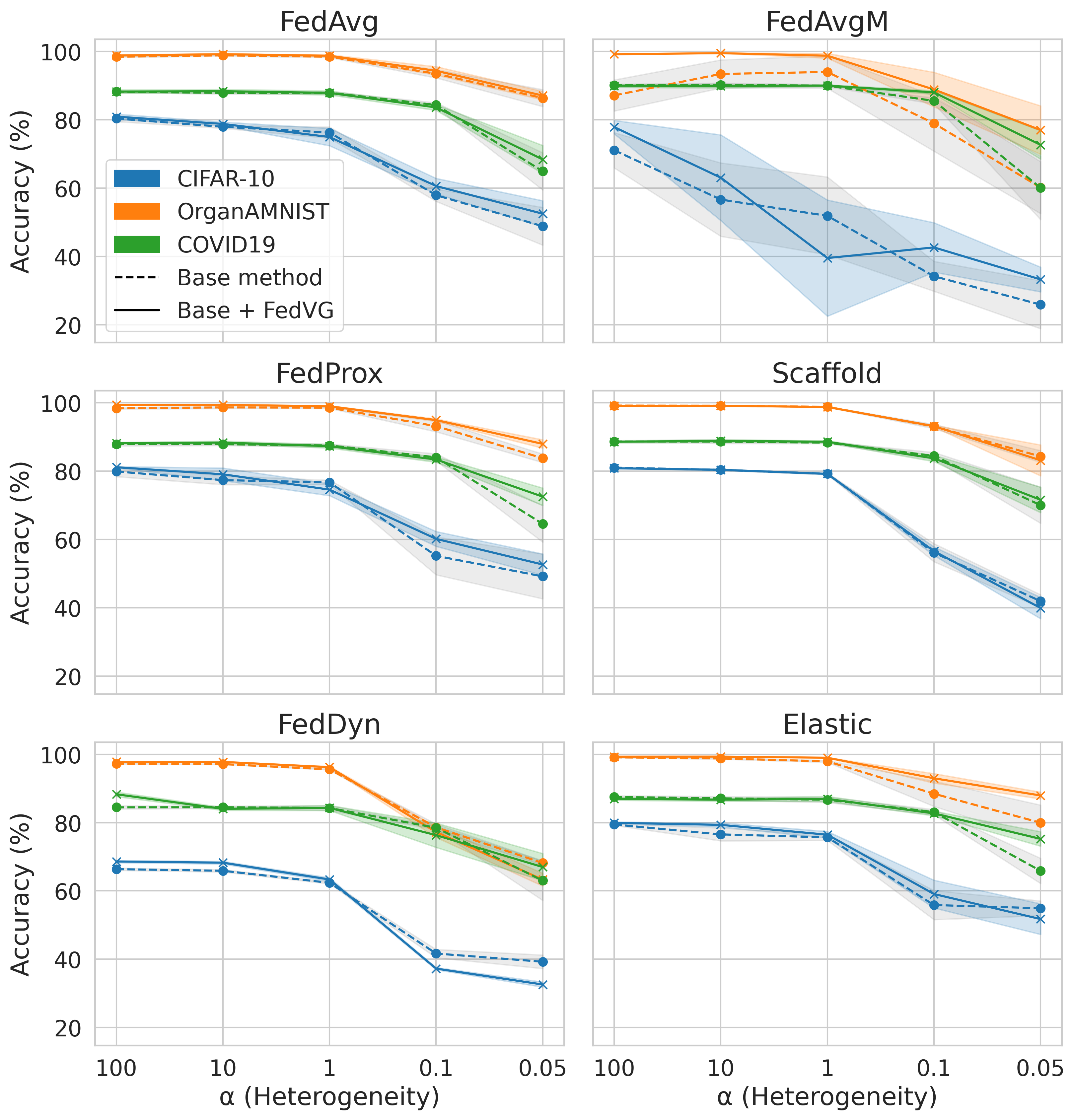}
\caption{
FedVG integration performance across CIFAR-10, OrganAMNIST, and COVID19 as $\alpha \rightarrow{} 0$.
Dashed lines denote the base methods, and solid lines their FedVG-enhanced counterparts. Shaded gray areas show the standard deviation of base methods, while colored regions reflect the variability of enhanced methods.
}
\label{fig:res_integration}
\end{figure}

\vspace{-3pt}
\subsection{Integrating FedVG with FL Algorithms}
\vspace{-3pt}
One key strength of FedVG is its modularity, allowing it to be easily integrated into existing FL algorithms by augmenting their aggregation strategies with gradient-based validation guidance. 
To evaluate this capability, we conduct experiments where FedVG is integrated into all six baseline methods, using their local update rules and regularization schemes while incorporating FedVG's validation-gradient-based weighting for server-side aggregation. 
For FedAvg, we take the mean of FedAvg and FedVG's weights during aggregation. As shown in Fig. ~\ref{fig:res_integration}, the hybrid approach of FedAvg + FedVG consistently improves upon vanilla FedAvg across all datasets and higher heterogeneity levels. 
Similar performance gains are observed when integrating FedProx with FedVG. 
While FedAvgM performs strongly in certain scenarios, it is notably sensitive to hyperparameter tuning; nevertheless, combining it with FedVG yields improvements.  
In contrast, Scaffold and FedDyn show more mixed results when integrated with FedVG. 
Finally, Elastic stands out as a particularly strong baseline and often performs on par with FedVG. 
However, combining Elastic with FedVG further improves performance on certain datasets 
, demonstrating that even top-performing methods can benefit from validation-guided aggregation.

These findings indicate that FedVG can serve as a complementary module, enhancing a wide range of FL strategies.
By leveraging validation gradients as a signal for server-side aggregation, FedVG can improve performance without requiring changes to client-side optimization.

\vspace{-4pt}
\subsection{External public dataset as global validation set}
\vspace{-4pt}
\label{sec:realistic}

\begin{figure}[t]
\centering
\includegraphics[width=1\columnwidth]{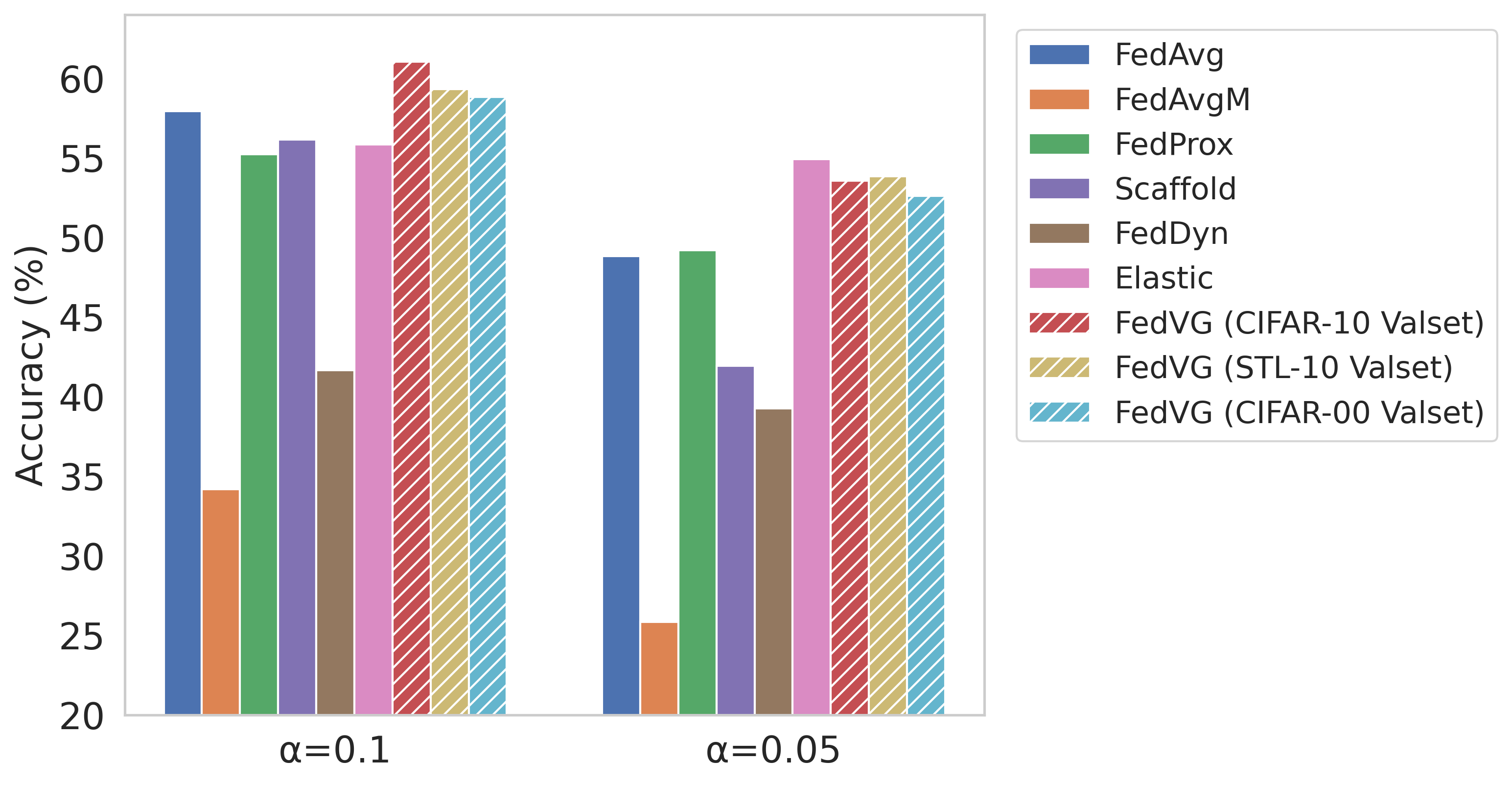}
\caption{Performance of baselines and FedVG (hatched) using an external global validation set at $\alpha=0.1$ and $0.05$.}
\label{fig:results_realistic}
\end{figure}

We implemented FedVG on CIFAR-10 using external public datasets (STL-10 \cite{coates2011analysis} and CIFAR-100 \cite{krizhevsky2009learning}) selected based on their imaging context and class overlaps. Nine classes are common between STL-10 and CIFAR-10 and three classes were mapped between CIFAR-10 and CIFAR-100. Images were randomly sampled from these datasets to form a small validation set of sizes 3650 and 4950 from CIFAR-100 and STL-10, respectively. Here, we assess the robustness and generalization of FedVG when aggregated using a validation set that differs from the training data distribution. Fig. ~\ref{fig:results_realistic} shows accuracies of baselines and FedVG variants under high heterogeneity ($\alpha = 0.1, 0.05$). FedVG methods consistently achieves high accuracies. FedVG achieved accuracies of 61.06\% and 53.58\%, while the variant validated on STL-10 achieved 59.32\% and 53.85\%  and the variant validated on CIFAR-100 achieved 58.83\% and 52.62\% at $\alpha = 0.1$ and $0.05$, respectively.
These results
demonstrate that even with an external validation set and distribution shifts, FedVG maintains superior performance.

\vspace{-4pt}
\subsection{Ablation Analysis}
\vspace{-4pt}
In this section, we investigate the impact of various aspects of FedVG, including the global validation set, norm type, and aggregation granularity.
Experiments are primarily conducted on CIFAR-10 using a ResNet-18 architecture unless stated otherwise.

\begin{figure}[t]
\centering
\includegraphics[width=1\columnwidth]{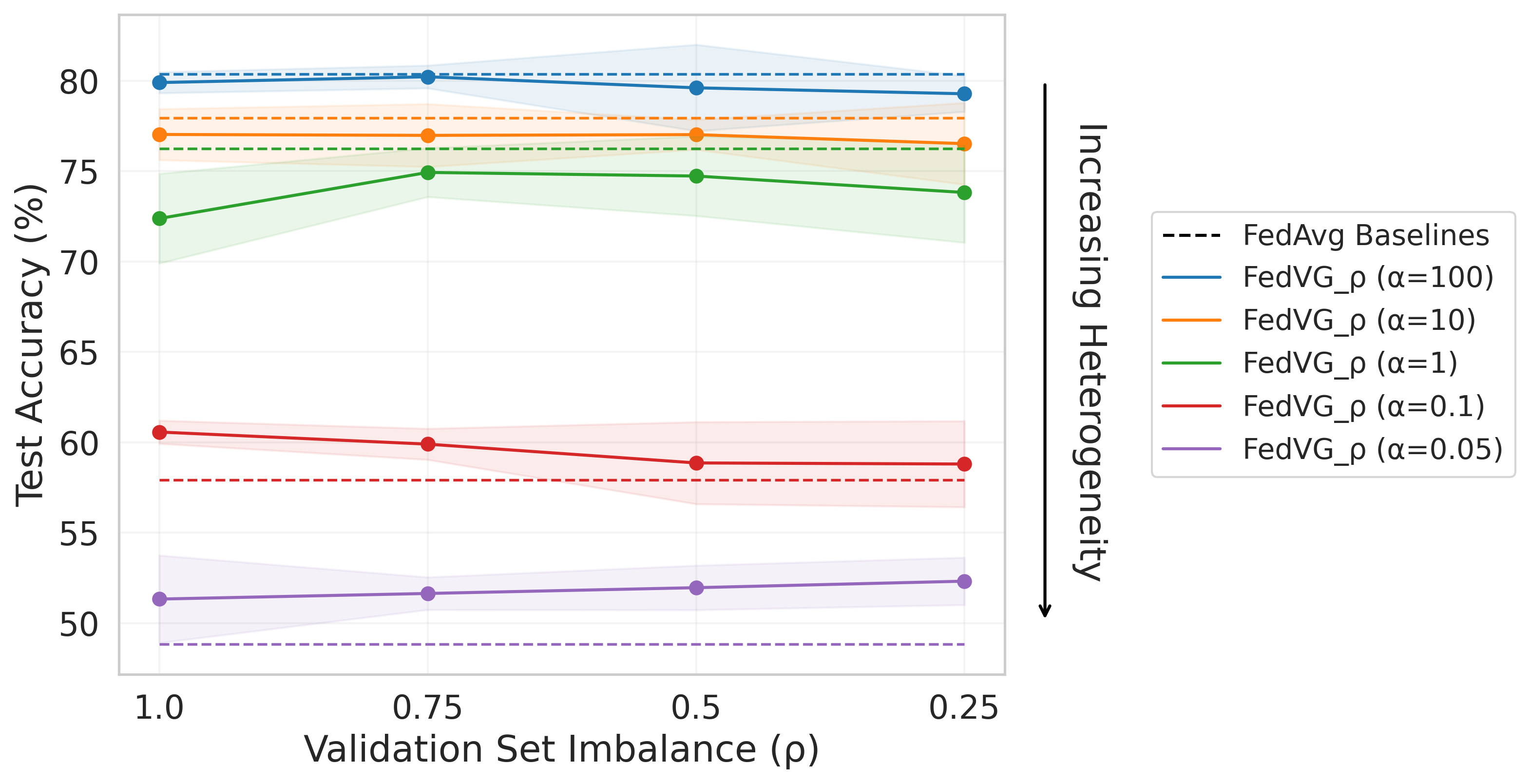}
\caption{FedVG performance across validation class imbalance ratios, with lower $\rho$ indicating greater class imbalance.
}
\label{fig:res_ablation_val_dist}
\end{figure}

\vspace{-4pt}
\paragraph{Analysis of class imbalance in global validation set $D_{\mathrm{val}}$:}
\vspace{-4pt}
In a real-world setting, client data distributions are unknown, and the global validation set may differ from the client distributions, often in terms of class distribution. 
Thus, to evaluate the robustness of FedVG under such 
distribution shift, we analyze its performance with class-imbalanced global validation sets. 
We define $\rho \in (0, 1]$ as the class imbalance ratio, where $\rho = 1$ indicates perfect balance. We analyze the FedVG's performance as $\rho \xrightarrow{}0$. 
We provide the specifics of the class imbalance settings in the Appendix, and the corresponding results are presented in Fig. \ref{fig:res_ablation_val_dist}.
As shown, FedVG consistently outperforms FedAvg even under high class imbalance, demonstrating the robustness of our approach to varying configurations of $D_{val}$.

\vspace{-4pt}
\paragraph{Evaluation of norm type}
\vspace{-4pt}
\label{sec:norm_types}
Here, we study the effect of different norm types on the calculation of $\hat{s}_k$. Specifically, compared to L1 used in FedVG, we analyze L2 and 
spectral norms \cite{spectral-norm}. Following prior work \cite{li2020federated} which used client drift as a penalty term, we also 
examine
the effect of $\hat{s}_k = || \theta_g - \theta_K ||$, referred to as \textit{delta norm}.

\begin{figure}[t]
\centering
\includegraphics[width=1\columnwidth]{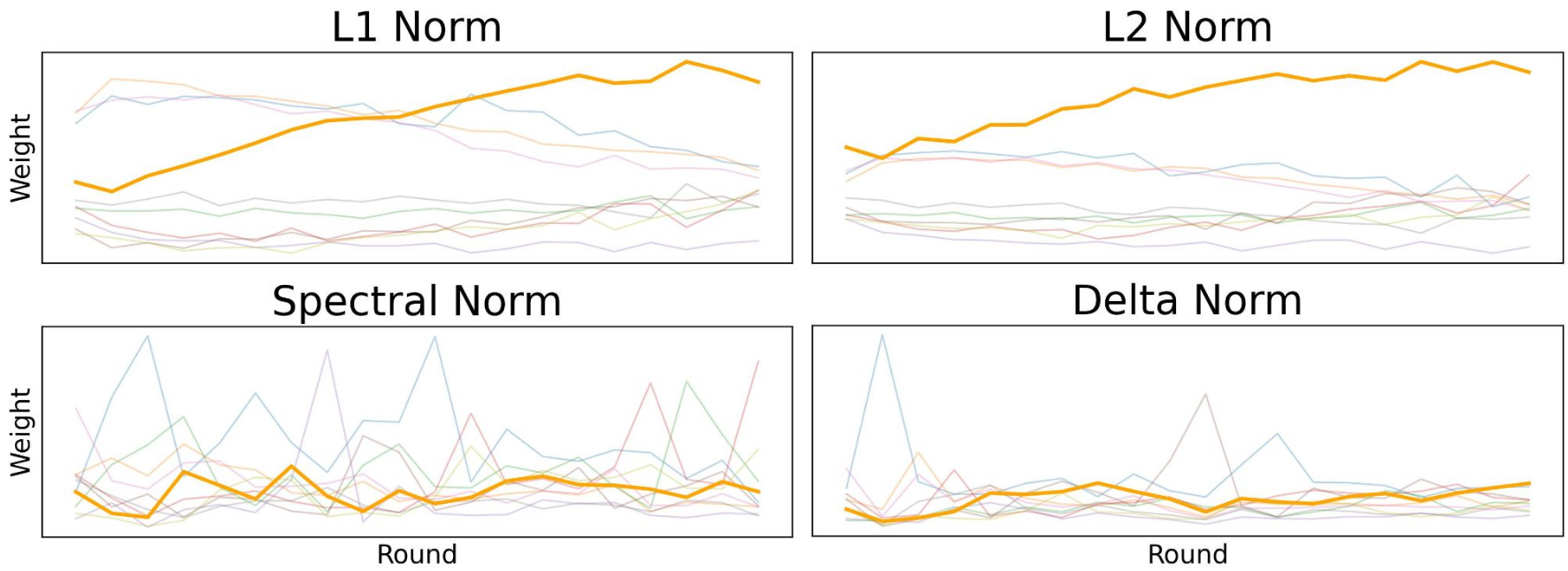}
\caption{Client weights for varying norms during weight calculation, where the \textbf{bold} line corresponds to the balanced client.}
\label{fig:norm_abl}
\end{figure}

To ensure a fair comparison, we construct a federated environment with 9 heterogeneous clients ($\alpha = 0.05$), each with $n_k = M$ training samples.
We then add a 10th, balanced, client, $c^*$.
Intuitively, $c^*$ should learn the highest-quality features, relative to the other clients in the environment, and should
therefore receive the highest aggregation weight. 
To evaluate the impact of 
each norm approach, we plot the client weights 
over 200 rounds of federated training in Fig. \ref{fig:norm_abl}.
As shown, both L1 and L2 norm methods are effective at properly assigning high weights to $c^*$, whereas the spectral and delta norms do not appear to recognize $c^*$ as the most prominent model. Quantitatively, FedVG attained accuracies of 70.36\%, 70.43\%, 68.50\%, 69.99\% for L1, L2, spectral, and delta norms respectively. Further details on this experiment are provided in the Appendix.

\begin{figure}[t]
\centering
\includegraphics[width=0.9\columnwidth]{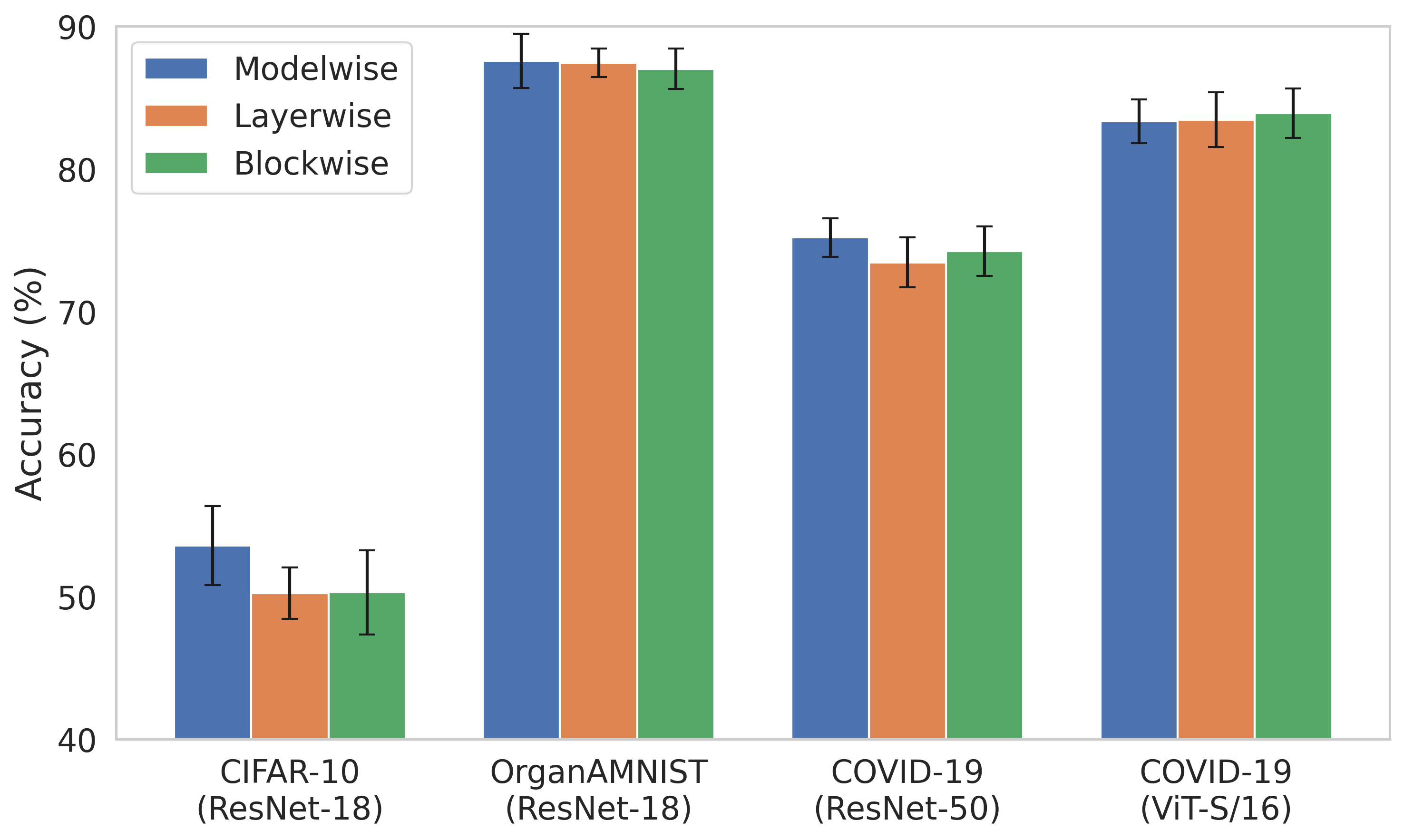}
\caption{Comparison of FedVG aggregation strategies
at $\alpha = 0.05$. Each bar shows the mean accuracy ± std over 5 runs.}
\label{fig:res_ablation_grabularity}
\end{figure}

\vspace{-5pt}
\paragraph{Aggregation granularity}
\vspace{-4pt}
We analyze the impact of aggregation granularity by comparing modelwise, layerwise, and blockwise aggregation strategies to understand how the level of parameter grouping influences FedVG's performances.
In the layerwise variant, aggregation weights, $s_{k}^{(l)}$ are computed for each individual layer $l$.
In the blockwise variant, weights $s_{k}^{(b)}$ are computed for blocks of layers $b$, where each block corresponds to a logical unit within the model architecture. In ResNet, each residual block is treated as a separate block, and in ViT, each transformer encoder block is treated as a block, with the classification heads considered separately.
In its default form, FedVG performs modelwise aggregation by calculating $s_k$ based on the gradient norms averaged for the entire local model.

We present results for $\alpha=0.05$ on some dataset-model configuration in Fig. ~\ref{fig:res_ablation_grabularity}, with full results available in Appendix.
Our results reveal mixed trends.  Modelwise FedVG performs best on some datasets, such as CIFAR-10 and OrganAMNIST. 
However, on others, particularly COVID19 and DermaMNIST, layerwise and blockwise strategies on ResNet-50 perform competitively or even slightly better under high heterogeneity. 
For ViT models, no single granularity level
consistently outperforms the others; different strategies excel depending on the setup.
These observations suggest that the optimal aggregation granularity depends on factors such as the model architecture, dataset characteristics, and data heterogeneity. 
While modelwise aggregation is often strong and simple, finer-grained variants like layerwise or blockwise can offer advantages in certain scenarios.



%% file: sec/6_conclusion.tex
\section{Conclusion}
\vspace{-5pt}

In this work, we introduce FedVG, a novel gradient-based FL aggregation method, which utilizes a global validation set to assign higher weights to clients with 
smaller
validation gradients. This method promotes generalization in highly heterogeneous environments, which are challenging for FL algorithms, by placing more emphasis on clients that learn high-quality features. Experiments across 
five publicly available datasets,
diverse model architectures, and varying levels of heterogeneity consistently demonstrate FedVG's robustness and superior performance compared to state-of-the-art FL approaches. 
Furthermore, FedVG’s modularity allows it to be integrated into existing FL strategies to improve their performance without requiring changes to client-side optimization.



A limitation of our study can be viewed in the potential sensitivity of FedVG to overlaps between the global validation set and certain clients’ data. While FedVG is intentionally designed not to bias contributions based on class distribution mismatches, other forms of similarity, such as shared domains, correlated samples, or preprocessing pipelines, may still introduce unintended imbalance. A deeper investigation of these overlap effects, and strategies to further strengthen fairness and generalization, is left to future work. Finally, although computing validation gradients introduces additional overhead, this burden is placed entirely on the server, imposing no extra cost on resource-constrained clients.

\vspace{-4pt}
\subsection*{Acknowledgment}
\vspace{-5pt}
This material is based upon work supported by the National Science Foundation (NSF) under Grant No. 2119654. Any opinions, findings, conclusions, or recommendations expressed in this material are those of the author(s) and do not necessarily reflect the views of the NSF. We also recognize the computational resources provided by the WVU Research Computing Dolly Sods HPC cluster, which is funded in part by NSF OAC-2117575.


%% file: sec/X_suppl.tex
\clearpage
\setcounter{page}{1}

\setcounter{section}{0}

\renewcommand{\thesection}{A.\arabic{section}}

\renewcommand{\thesubsection}{\thesection.\arabic{subsection}}

\setcounter{figure}{0}
\renewcommand{\thefigure}{A.\arabic{figure}}

\setcounter{table}{0}
\renewcommand{\thetable}{A.\arabic{table}}

\setcounter{equation}{0}
\renewcommand{\theequation}{E.\arabic{equation}}

\maketitlesupplementary

\section{Relationship Between Gradient Norms and Joint Fisher in the Zero-Limit}

Let the model conditional distribution be \(p_\theta(y\mid x)\).  
For the cross-entropy loss 
\begin{equation}
  \ell(\theta;x,y) = -\log p_\theta(y\mid x),  
\end{equation}

the gradient is
\begin{equation}
\nabla_\theta \ell(\theta;x,y)
= - \nabla_\theta \log p_\theta(y\mid x)
= - s(x,y;\theta),
\end{equation}

where \(s(x,y;\theta)\) is the score function.

The (population) Fisher Information Matrix is
\begin{equation}
    F(\theta)=\mathbb{E}[\,s s^\top\,],
\qquad
\end{equation}

and the Joint Fisher is defined element-wise by

\begin{equation}
    J(\theta)=\mathbb{E}[\,s\circ s\,].
\end{equation}

Using \(\|s\|_2^2 = s^\top s = \mathrm{tr}(s s^\top)\), we have
\begin{equation}
    \mathbb{E}[\|s\|_2^2] = \mathrm{tr}(F(\theta)) = \sum_j J_j(\theta).
\end{equation}

Because \(\|s\|_2^2 \ge 0\), the equality
\begin{equation}
    \mathrm{tr}(F(\theta)) = 0 
\quad\Longleftrightarrow\quad 
\mathbb{E}[\|s\|_2^2]=0.
\end{equation}
holds if and only if 
$\|s(x,y;\theta)\|_2 = 0$ almost surely.

Consequently,  
\begin{equation}
\nabla_\theta \ell(\theta;x,y)=0
\qquad\text{and}\qquad
J(\theta)=0.
\end{equation}

Thus, for cross-entropy, vanishing Joint Fisher is equivalent to vanishing gradient norms, showing that both quantities measure the same “zero-sensitivity” or "flatness" condition of the model i.e., the parameters no longer respond to perturbations in the data.

\section{Behaviors of Layers in an FL Setting}

\begin{figure*}[ht]
\centering
\includegraphics[width=0.9\textwidth]{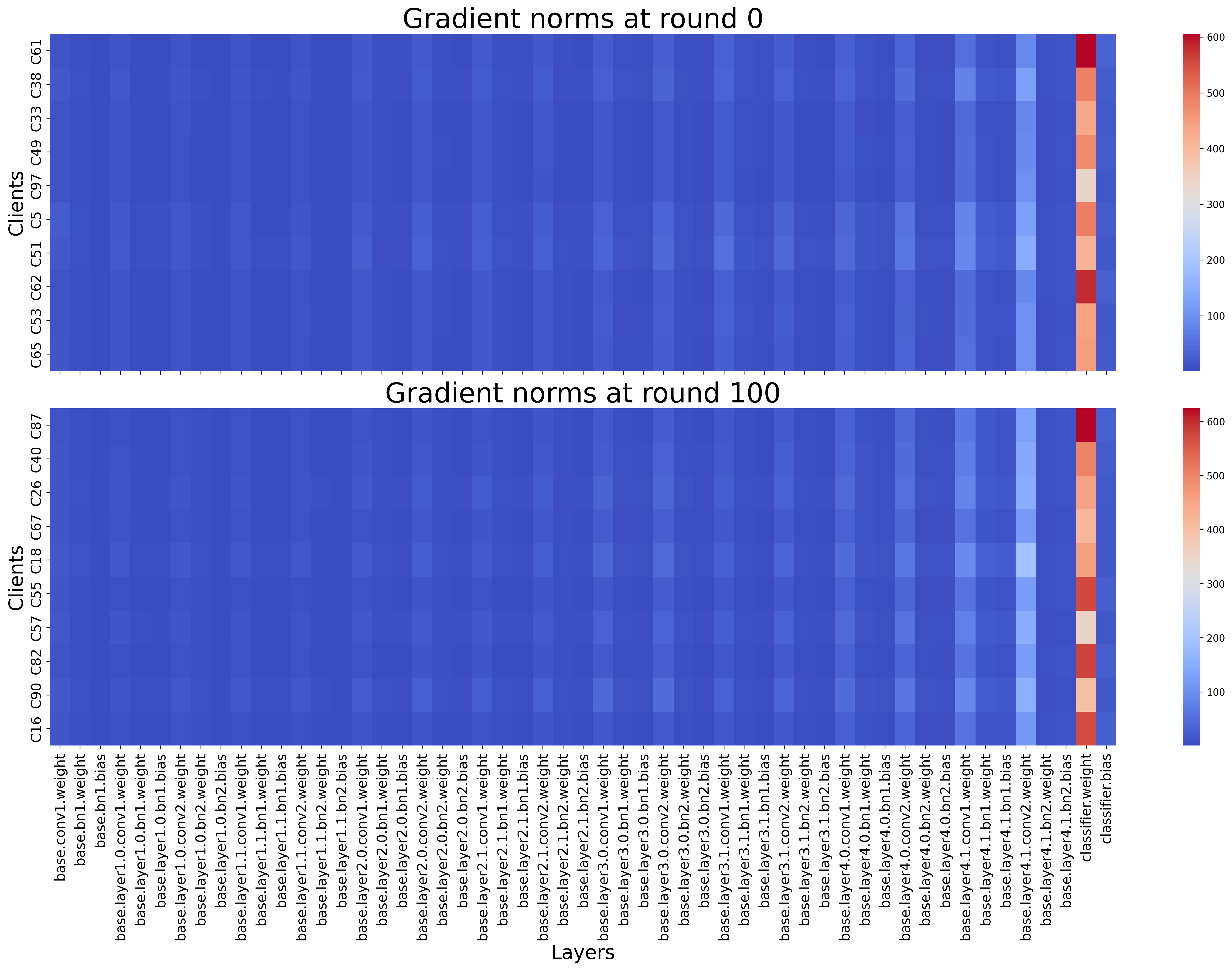}
\caption{Heatmaps of per-client gradient norms across network layers for two federated learning rounds. Only a subset of layers is presented for clarity.}
\label{fig:layer_norms}
\end{figure*}

Fig. ~\ref{fig:layer_norms} presents heatmaps of client-wise gradient norms across a selected subset of network layers at two training rounds. We observe that the later layers consistently exhibit larger gradient norms, reflecting their greater sensitivity to client-specific updates. In contrast, earlier layers show smaller and more uniform magnitudes. The variation across layers highlights distinct gradient behaviors within the network.

\section{Additional Details of Experimental Settings}
This section presents additional details on the hyperparameter configurations used in our experiments.

\subsection{Details of Hyperparameter Configurations}

Table ~\ref{tab:fl_hyperparameters} summarizes the hyperparameter configurations used in our federated learning experiments for each dataset. While certain parameters, such as the optimizer (SGD), learning rate (0.01), number of communication rounds (200), and local epochs (5), are consistent across datasets, others are tailored to dataset characteristics. While the number of training rounds is fixed at 200, model selection is performed to identify the best round and corresponding model based on performance. Table~\ref{tab:baseline_hyperparameters} lists the baseline-specific hyperparameters used in our experiments.

\begin{table*}[ht]
\centering
\caption{Hyperparameter configurations used for different datasets in the federated learning experiments.}
\begin{tabular}{cccccc}
\toprule
 & CIFAR-10 & OrganAMNIST & Tiny ImageNet & COVID19 & DermaMNIST \\
\midrule
Number of Clients & 100 & 100 & 100 & 20 & 25 \\
Join Ratio & 0.1 & 0.1 & 0.1 & 0.25 & 0.2 \\
Input Channels & 3 & 1 & 3 & 3 & 3 \\
Image Size & 32$\times$32 & 28$\times$28 & 64$\times$64 & 244$\times$224 & 224$\times$224 \\
Batch Size & 32 & 32 & 128 & 32 & 32 \\
Optimizer & SGD & SGD & SGD & SGD & SGD \\
Learning Rate & 0.01 & 0.01 & 0.01 & 0.01 & 0.01 \\
Momentum & 0 & 0 & 0.9 & 0 & 0 \\
Number of Rounds & 200 & 200 & 200 & 200 & 200 \\
Local Epochs & 5 & 5 & 5 & 5 & 5 \\
\bottomrule
\end{tabular}
\label{tab:fl_hyperparameters}
\end{table*}

\begin{table}[ht]
\centering
\caption{Baseline-specific hyperparameters used in our experiments.}
\begin{tabular}{ccc}
\toprule
\textbf{Baseline} & \textbf{Hyperparameter} & \textbf{Value} \\
\midrule
FedAvgM & server momentum & 0.9 \\
\midrule
FedProx & $\mu$ & 0.01 \\
\midrule
Scaffold & global lr & 1.0 \\
\midrule
\multirow{2}{*}{FedDyn} & $\alpha$ & 0.1 \\
                        & max\_grad\_norm & 10 \\
\midrule
\multirow{3}{*}{Elastic} & sample\_ratio & 0.3 \\
                         & $\tau$ & 0.5 \\
                         & $\mu$ & 0.95 \\
\bottomrule
\end{tabular}
\label{tab:baseline_hyperparameters}
\end{table}


\subsection{Examples of Non-IID Client Distribution}
\begin{figure*}[ht]
     \centering
     \begin{subfigure}[t]{0.19\textwidth}
         \centering
         \includegraphics[width=\textwidth]{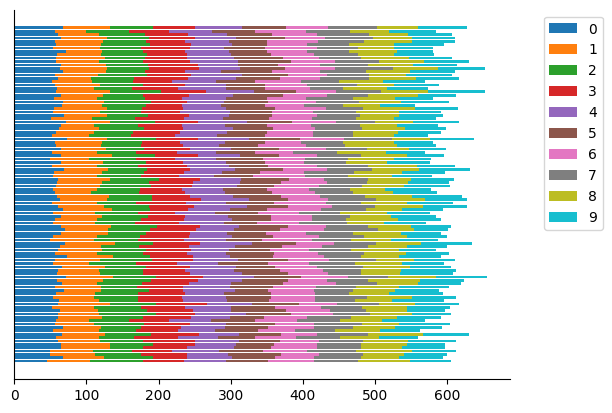}
         \caption{$\alpha = 100$}
     \end{subfigure}
     \hfill
     \begin{subfigure}[t]{0.19\textwidth}
         \centering
         \includegraphics[width=\textwidth]{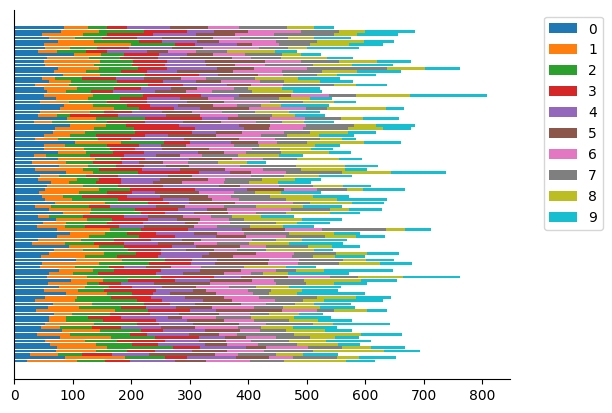}
         \caption{$\alpha = 10$}
     \end{subfigure}
     \hfill
     \begin{subfigure}[t]{0.19\textwidth}
         \centering
         \includegraphics[width=\textwidth]{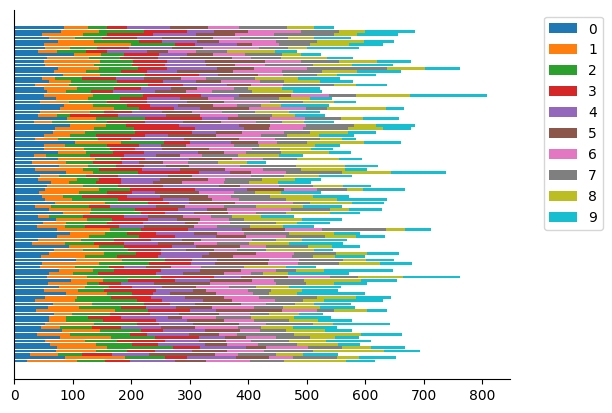}
         \caption{$\alpha = 1$}
     \end{subfigure}
     \hfill
     \begin{subfigure}[t]{0.19\textwidth}
         \centering
         \includegraphics[width=\textwidth]{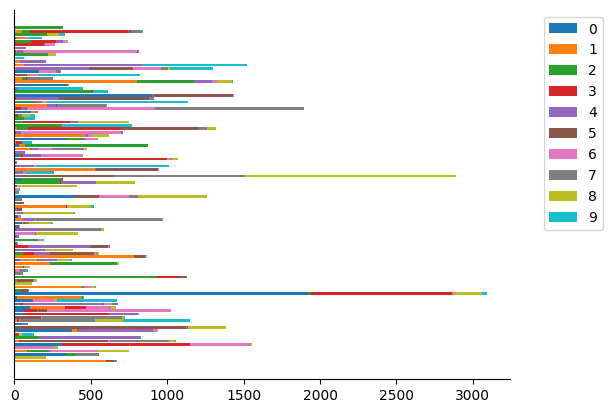}
         \caption{$\alpha = 0.1$}
     \end{subfigure}
     \hfill
     \begin{subfigure}[t]{0.19\textwidth}
         \centering
         \includegraphics[width=\textwidth]{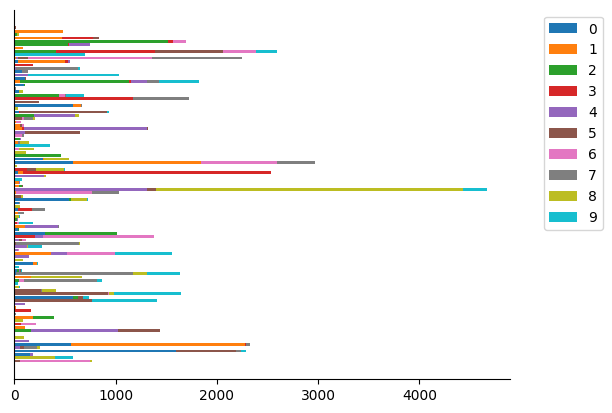}
         \caption{$\alpha = 0.05$}
     \end{subfigure}
        \caption{CIFAR-10 class distributions at different heterogeneity levels, parameterized by $\alpha$ in the Dirichlet distribution. Lower $\alpha$ indicates greater heterogeneity. Each row corresponds to a client, and colors represent different classes.}
        \label{fig:cifar10_non_iid}
\end{figure*}

\begin{figure*}[ht]
     \centering
     \begin{subfigure}[t]{0.19\textwidth}
         \centering
         \includegraphics[width=\textwidth]{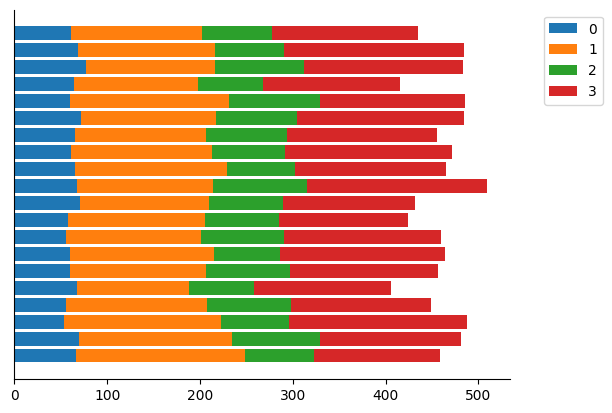}
         \caption{$\alpha = 100$}
     \end{subfigure}
     \hfill
     \begin{subfigure}[t]{0.19\textwidth}
         \centering
         \includegraphics[width=\textwidth]{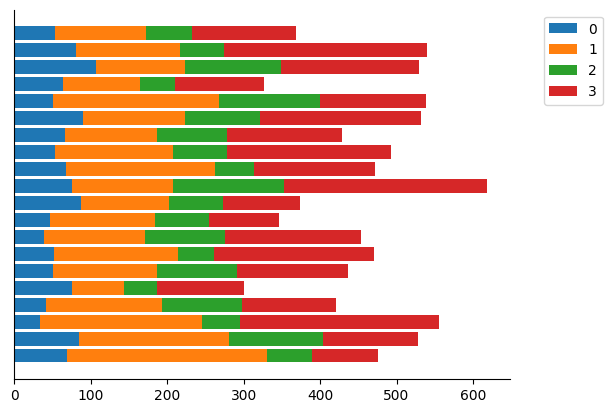}
         \caption{$\alpha = 10$}
     \end{subfigure}
     \hfill
     \begin{subfigure}[t]{0.19\textwidth}
         \centering
         \includegraphics[width=\textwidth]{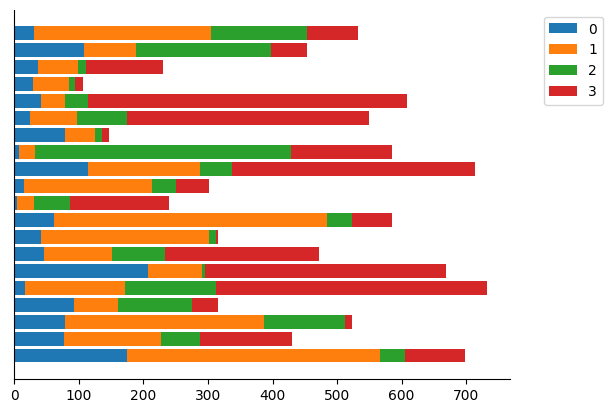}
         \caption{$\alpha = 1$}
     \end{subfigure}
     \hfill
     \begin{subfigure}[t]{0.19\textwidth}
         \centering
         \includegraphics[width=\textwidth]{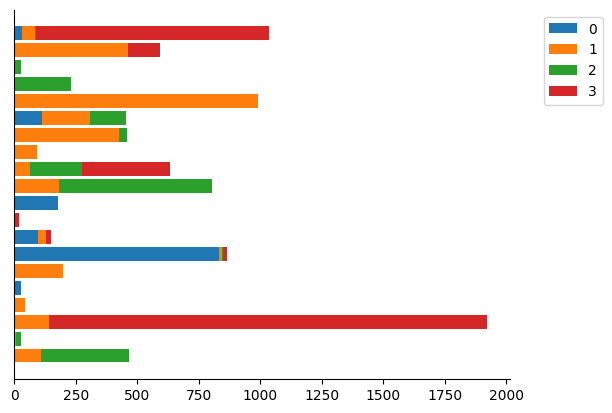}
         \caption{$\alpha = 0.1$}
     \end{subfigure}
     \hfill
     \begin{subfigure}[t]{0.19\textwidth}
         \centering
         \includegraphics[width=\textwidth]{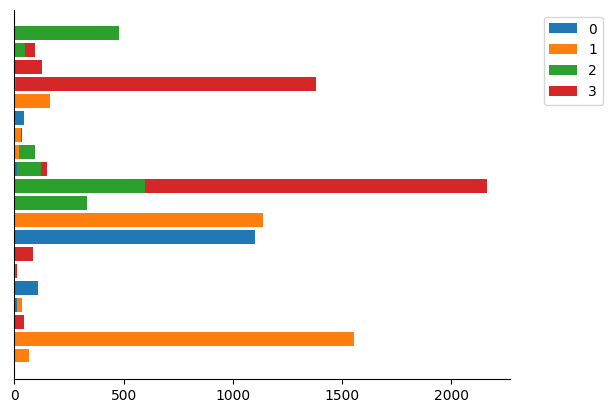}
         \caption{$\alpha = 0.05$}
     \end{subfigure}
        \caption{COVID19 class distributions at different heterogeneity levels, parameterized by $\alpha$ in the Dirichlet distribution. Lower $\alpha$ indicates greater heterogeneity. Each row corresponds to a client, and colors represent different classes.}
        \label{fig:covid_non_iid}
\end{figure*}

Figure~\ref{fig:cifar10_non_iid} illustrates the CIFAR-10 class distributions across clients for different levels of data heterogeneity, controlled by the Dirichlet concentration parameter $\alpha$. Lower values of 
$\alpha$ corresponds to more skewed label distributions, resulting in higher non-IID heterogeneity across clients. Figure~\ref{fig:covid_non_iid} illustrates the COVID19 class distributions across clients for different values of $\alpha$.

\section{Additional Results of FedVG Performance}

In this section, we provide detailed numerical results of various federated learning algorithms in tabular form to facilitate clearer comparisons.

\subsection{Performance on ResNet Model Architectures}

\begin{figure}[ht]
\centering
\includegraphics[width=0.4\textwidth]{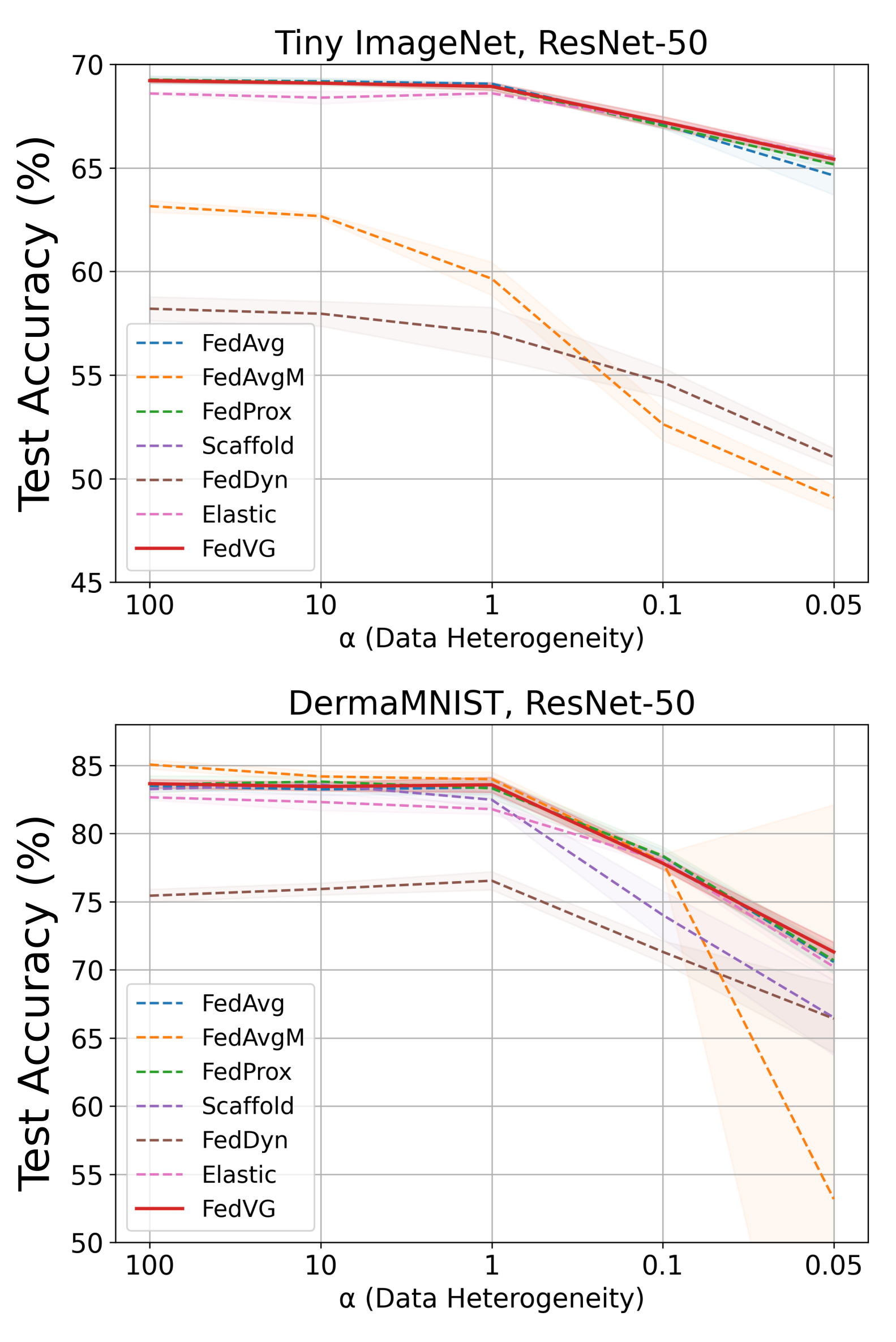}
\caption{FL algorithm performance for TinyImageNet and DermaMNIST datasets on the ResNet-50 model as $\alpha \rightarrow{} 0$. Shaded areas indicate standard deviations.}
\label{fig:res_resnet_remaining}
\end{figure}

Fig. ~\ref{fig:res_resnet_remaining} shows the performances of FedVG and baseline algorithms for Tiny-ImageNet and DermaMNIST datasets on ResNet-50 model.
Table ~\ref{tab:fl_results_resnet} summarizes the overall performance of various federated learning methods on ResNet-based models across different dataset-model combinations and degrees of data heterogeneity.

\begin{table*}[t]
\centering
\caption{Test accuracy (\%) ± standard deviation of various federated learning methods on CIFAR-10, OrganAMNIST, Tiny ImageNet, COVID19, and DermaMNIST datasets across different data heterogeneity levels (Dirichlet $\alpha$). \textbf{Bold} values indicate best accuracies and \underline{underlined} values indicate second best accuracies.}
\begin{tabular}{l|ccccc}
\toprule
Methods & \textbf{$\alpha=100$} & \textbf{$\alpha=10$} & \textbf{$\alpha=1$} & \textbf{$\alpha=0.1$} & \textbf{$\alpha=0.05$} \\
\midrule
\multicolumn{6}{c}{\textit{CIFAR-10 / ResNet-18}} \\
\midrule
FedAvg & 80.37 ± 0.89 & 77.95 ± 0.43 & 76.26 ± 1.65 & \underline{57.93 ± 1.82} & 48.83 ± 5.53 \\
FedAvgM & 71.09 ± 5.16 & 56.64 ± 10.75 & 51.87 ± 11.40 & 34.18 ± 4.38 & 25.87 ± 7.04 \\
FedProx & 79.92 ± 1.50 & 77.39 ± 1.32 & \underline{76.73 ± 0.63} & 55.22 ± 5.56 & 49.19 ± 6.55 \\
Scaffold & \textbf{81.05 ± 0.16} & \textbf{80.39 ± 0.15} & \textbf{79.25 ± 0.23} & 56.15 ± 2.65 & 41.94 ± 1.91 \\
FedDyn & 66.37 ± 0.33 & 65.90 ± 0.36 & 62.37 ± 0.29 & 41.66 ± 1.22 & 39.26 ± 1.99 \\
Elastic & 79.45 ± 0.27 & 76.55 ± 1.88 & 75.69 ± 0.89 & 55.86 ± 4.26 & \textbf{54.92 ± 2.16} \\
FedVG (Ours) & \underline{80.74 ± 0.79} & \underline{79.65 ± 0.52} & 75.69 ± 2.27 & \textbf{61.06 ± 0.34} & \underline{53.58 ± 2.78} \\
\midrule
\multicolumn{6}{c}{\textit{OrganAMNIST / ResNet-18}} \\
\midrule
FedAvg & 98.46 ± 0.16 & 98.87 ± 0.21 & 98.47 ± 0.24 & \underline{93.50 ± 1.22} & \underline{86.37 ± 2.49} \\
FedAvgM & 87.14 ± 4.56 & 93.43 ± 4.13 & 94.00 ± 4.76 & 78.95 ± 8.14 & 60.15 ± 7.61 \\
FedProx & 98.44 ± 0.21 & 98.68 ± 0.32 & 98.60 ± 0.20 & 93.19 ± 1.63 & 83.80 ± 1.25 \\
Scaffold & \underline{99.22 ± 0.06} & \underline{99.17 ± 0.06} & \underline{98.85 ± 0.06} & 93.11 ± 0.62 & 84.32 ± 1.72 \\
FedDyn & 97.26 ± 0.17 & 97.13 ± 0.07 & 95.60 ± 0.15 & 78.61 ± 1.01 & 68.17 ± 0.62 \\
Elastic & 99.12 ± 0.11 & 98.78 ± 0.18 & 97.94 ± 0.27 & 88.46 ± 3.70 & 79.96 ± 5.20 \\
FedVG (Ours) & \textbf{99.41 ± 0.08} & \textbf{99.42 ± 0.02} & \textbf{99.12 ± 0.10} & \textbf{94.72 ± 0.68} & \textbf{87.57 ± 1.91} \\
\midrule
\multicolumn{6}{c}{\textit{Tiny ImageNet / ResNet-50}} \\
\midrule
FedAvg & \underline{69.25 ± 0.19} & \textbf{69.18 ± 0.16} & \textbf{69.06 ± 0.08} & 67.08 ± 0.18 & 64.63 ± 0.93 \\
FedAvgM & 63.15 ± 0.29 & 62.67 ± 0.14 & 59.65 ± 0.80 & 52.63 ± 0.79 & 49.08 ± 0.61 \\
FedProx & \textbf{69.26 ± 0.14} & \underline{69.13 ± 0.17} & \underline{68.91 ± 0.16} & 67.04 ± 0.16 & 65.17 ± 0.22 \\
Scaffold & 40.53 ± 2.33 & 40.55 ± 2.53 & 37.85 ± 3.29 & 34.40 ± 5.73 & 29.23 ± 3.54 \\
FedDyn & 58.20 ± 0.58 & 57.96 ± 0.60 & 57.05 ± 1.22 & 54.65 ± 0.70 & 51.03 ± 0.43 \\
Elastic & 68.59 ± 0.05 & 68.39 ± 0.30 & 68.60 ± 0.11 & \textbf{67.23 ± 0.04} & \textbf{65.50 ± 0.40} \\
FedVG (Ours) & 69.21 ± 0.10 & 69.09 ± 0.05 & 68.93 ± 0.18 & \underline{67.21 ± 0.27} & \underline{65.42 ± 0.18} \\
\midrule
\multicolumn{6}{c}{\textit{COVID19 / ResNet-50}} \\
\midrule
FedAvg & 88.23 ± 0.23 & 87.88 ± 0.38 & 87.91 ± 0.37 & 84.34 ± 0.67 & 64.92 ± 5.45 \\
FedAvgM &  \textbf{90.16 ± 0.23} &  \textbf{90.23 ± 0.54} &  \textbf{90.00 ± 0.33} &  \textbf{85.54 ± 0.87} & 60.07 ± 9.31 \\
FedProx & 87.93 ± 0.55 & 87.91 ± 0.37 & 87.53 ± 0.47 & 84.06 ± 1.06 & 64.58 ± 5.34 \\
Scaffold & \underline{88.70 ± 0.30} & \underline{88.67 ± 0.55} & \underline{88.36 ± 0.26} & \underline{84.47 ± 1.06} & \underline{70.05 ± 5.29} \\
FedDyn & 84.48 ± 0.48 & 84.51 ± 0.33 & 84.28 ± 0.75 & 78.49 ± 0.48 & 63.01 ± 5.79 \\
Elastic & 87.50 ± 0.37 & 87.12 ± 0.57 & 86.69 ± 0.78 & 83.09 ± 0.79 & 65.87 ± 3.69 \\
FedVG (Ours) & 88.30 ± 0.66 & 87.83 ± 0.39 & 87.40 ± 0.74 & 83.10 ± 0.54 & \textbf{75.18 ± 1.36} \\
\midrule
\multicolumn{6}{c}{\textit{DermaMNIST / ResNet-50}} \\
\midrule
FedAvg & 83.49 ± 0.20 & 83.24 ± 0.42 & 83.38 ± 0.51 & \underline{78.32 ± 0.50} & 70.58 ± 0.87 \\
FedAvgM & \textbf{85.06 ± 0.33} & \textbf{84.19 ± 0.38} & \textbf{83.99 ± 0.52} & 77.92 ± 0.56 & 53.18 ± 28.91 \\
FedProx & 83.61 ± 0.62 & \underline{83.81 ± 0.39} & 83.32 ± 0.77 & \textbf{78.37 ± 0.66} & \underline{70.70 ± 0.87} \\
Scaffold & 83.26 ± 0.21 & 83.59 ± 0.20 & 82.48 ± 0.59 & 74.02 ± 1.77 & 66.50 ± 2.76 \\
FedDyn & 75.44 ± 0.47 & 75.93 ± 0.44 & 76.54 ± 0.67 & 71.32 ± 0.81 & 66.42 ± 2.50 \\
Elastic & 82.66 ± 0.20 & 82.31 ± 0.62 & 81.79 ± 0.37 & 78.00 ± 0.38 & 70.22 ± 1.07 \\
FedVG (Ours) & \underline{83.66 ± 0.31} & 83.45 ± 0.31 & \underline{83.58 ± 0.55} & 77.82 ± 0.46 & \textbf{71.31 ± 0.71} \\
\bottomrule
\end{tabular}
\label{tab:fl_results_resnet}
\end{table*}

\subsection{Performance on Vision Transformer Model Architectures}

Table ~\ref{tab:fl_results_vit} summarizes the overall performance of various federated learning methods on ViT-based models across different dataset-model combinations and degrees of data heterogeneity.

\begin{table*}[t]
\centering
\caption{Test accuracy (\%) ± standard deviation of federated learning methods on COVID19 with ViT-S/16 and DermaMNIST with ViT-B/16 models across different data heterogeneity levels (Dirichlet $\alpha$). \textbf{Bold} values indicate best accuracies and \underline{underlined} values indicate second best accuracies.}
\begin{tabular}{l|ccccc}
\toprule
Methods & \textbf{$\alpha=100$} & \textbf{$\alpha=10$} & \textbf{$\alpha=1$} & \textbf{$\alpha=0.1$} & \textbf{$\alpha=0.05$} \\
\midrule
\multicolumn{6}{c}{\textit{COVID19 / ViT-S/16}} \\
\midrule
FedAvg & \underline{88.38 ± 0.73} & 89.07 ± 0.49 & \underline{89.63 ± 0.47} & 87.46 ± 0.72 & 82.30 ± 3.63 \\
FedAvgM & 88.01 ± 0.75 & 88.77 ± 0.50 & 89.14 ± 0.38 & 80.06 ± 8.58 & 51.66 ± 8.36 \\
FedProx & \textbf{88.40 ± 0.57} & \textbf{89.69 ± 0.12} & 89.32 ± 0.52 & \textbf{87.98 ± 1.06} & 81.27 ± 3.57 \\
Scaffold & 84.41 ± 4.02 & 84.80 ± 4.78 & 85.75 ± 4.33 & 68.33 ± 15.35 & 75.41 ± 9.81 \\
FedDyn & 77.61 ± 1.61 & 78.16 ± 1.64 & 76.72 ± 1.50 & 51.09 ± 8.63 & 42.42 ± 4.94 \\
Elastic & 88.37 ± 0.68 & \underline{89.13 ± 0.27} & \textbf{89.66 ± 0.80} & 87.04 ± 1.22 & \textbf{85.06 ± 1.05} \\
FedVG (Ours) & 88.29 ± 0.33 & 89.06 ± 0.29 & 89.26 ± 0.45 & \underline{87.47 ± 0.78} & \underline{83.34 ± 1.53} \\
\midrule
\multicolumn{6}{c}{\textit{DermaMNIST / ViT-B/16}} \\
\midrule
FedAvg & \underline{81.13 ± 0.27} & 81.31 ± 0.65 & 80.85 ± 0.50 & 77.79 ± 1.84 & 73.53 ± 2.40 \\
FedAvgM & 80.65 ± 0.73 & 81.08 ± 0.40 & 80.43 ± 1.39 & 71.74 ± 2.18 & 67.15 ± 0.32 \\
FedProx & 78.64 ± 6.28 & 80.85 ± 0.94 & 81.03 ± 0.31 & \underline{78.50 ± 1.02} & 73.29 ± 3.79 \\
Scaffold & \textbf{81.43 ± 0.48} & \underline{81.48 ± 0.42} & \underline{81.41 ± 0.38} & 76.44 ± 1.47 & 75.36 ± 2.68 \\
FedDyn & 67.85 ± 1.45 & 66.87 ± 0.06 & 66.34 ± 0.76 & 66.67 ± 0.56 & 66.98 ± 0.04 \\
Elastic & 80.96 ± 0.35 & 79.44 ± 3.27 & 81.03 ± 0.60 & \textbf{79.48 ± 0.67} & \underline{75.50 ± 1.41} \\
FedVG (Ours) & 80.46 ± 2.31 & \textbf{81.60 ± 0.26} & \textbf{81.46 ± 0.50} & 78.31 ± 0.99 & \textbf{76.20 ± 0.76} \\
\bottomrule
\end{tabular}
\label{tab:fl_results_vit}
\end{table*}

\subsection{FedVG Performances at High Heterogeneity ($\alpha=0.05$)}
\begin{figure}[t]
\centering
\includegraphics[width=1\columnwidth]{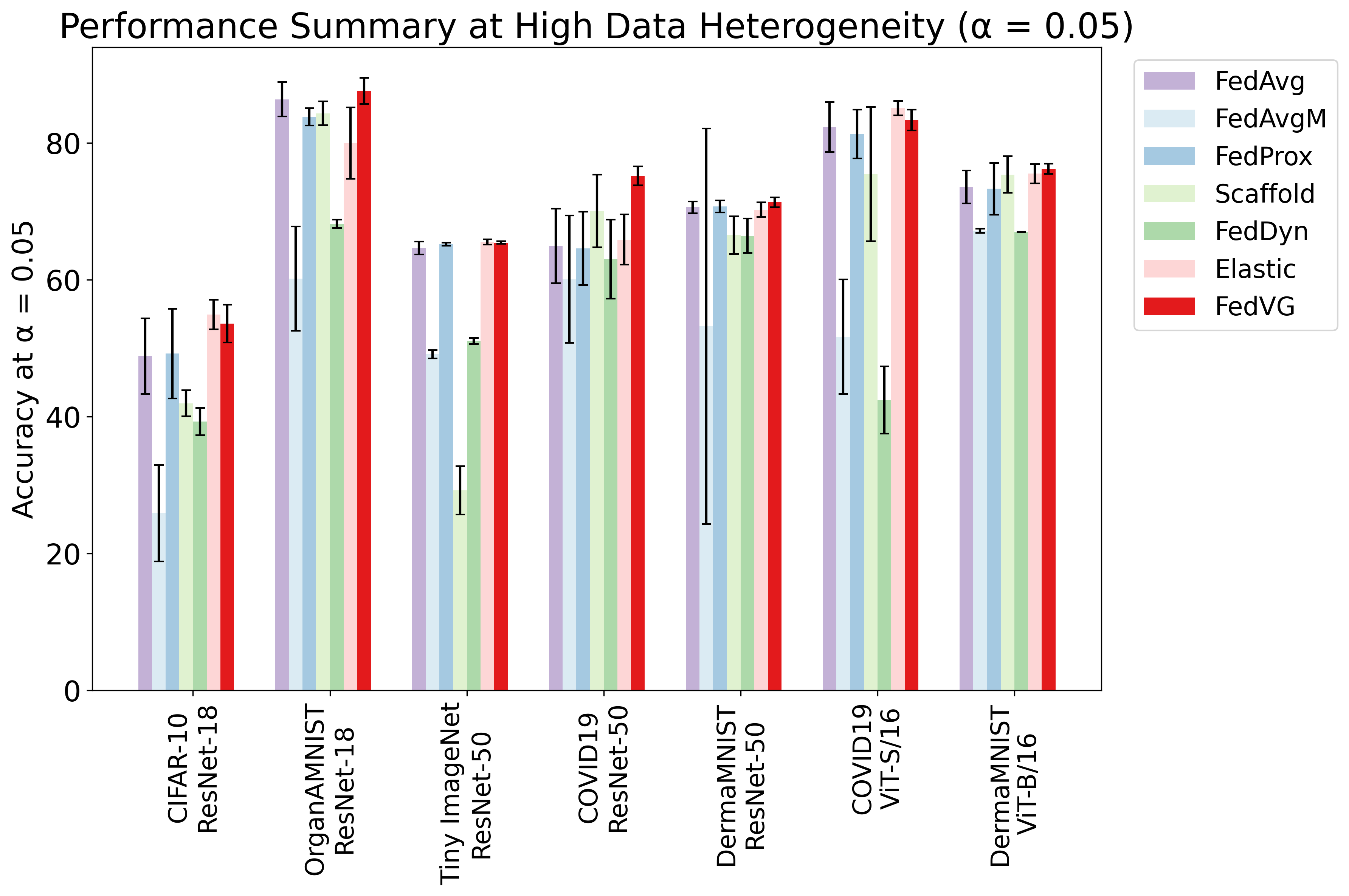}
\caption{Performance comparison of federated learning methods across datasets at high heterogeneity ($\alpha$ = 0.05). The error bars denote standard deviation across five runs. At high heterogeneity, FedVG maintains both strong average performance and low variance compared to other methods.}
\label{fig:results_high_het}
\end{figure}

Fig. ~\ref{fig:results_high_het} presents a grouped bar plot comparing the performance of various federated learning methods at a high level of data heterogeneity ($\alpha = 0.05$) across seven dataset–model combinations. FedVG consistently achieves high average accuracy while maintaining relatively low variance across different tasks, demonstrating its robustness to extreme non-IID settings. Elastic aggregation also emerges as a competitive baseline, achieving strong results in several cases, though with slightly higher variability in certain datasets.

\section{Validation set curation using public dataset}

\begin{table}[h]
\centering
\caption{Class mapping from STL-10 to CIFAR-10}
\begin{tabular}{c c c}
\toprule
\textbf{CIFAR-10} & \textbf{STL-10} & \textbf{Count} \\
\midrule
airplane & airplane & 550 \\
automobile & car & 550 \\
bird & bird & 550 \\
cat & cat & 550 \\
deer & deer & 550 \\
dog & dog & 550 \\
frog & --- & --- \\
horse & horse & 550 \\
ship & ship & 550 \\
truck & truck & 550 \\
\bottomrule
\label{tab:stl10_to_cifar10}
\end{tabular}
\end{table}

\begin{table}[h]
\centering
\caption{Class mapping from CIFAR-100 to CIFAR-10}
\begin{tabular}{c c c}
\toprule
\textbf{CIFAR-10} & \textbf{CIFAR-100} & \textbf{Count} \\
\midrule
airplane & --- & --- \\
automobile & bus, motorcycle & 531, 519 \\
bird & --- & --- \\
cat & leopard, lion, tiger & 520, 514, 516 \\
deer & --- & --- \\
dog & wolf, fox & 520, 530 \\
frog & --- & --- \\
horse & --- & --- \\
ship & --- & --- \\
truck & --- & --- \\
\bottomrule
\label{tab:cifar100_to_cifar10}
\end{tabular}
\end{table}

For the experiments in Section~\ref{fig:results_realistic}, we curated the validation set using CIFAR-100 and STL-10. Tables~\ref{tab:cifar100_to_cifar10} and~\ref{tab:stl10_to_cifar10} show the class mappings from CIFAR-100 and STL-10 to CIFAR-10, respectively. 
After mapping, we applied balanced random sampling to select a fixed number of examples per class from the external datasets, ensuring a representative validation set with uniform class coverage. A total of 3,650 samples from CIFAR-100 and 4,950 samples from STL-10 were selected.

\section{FedVG Integration with FL Algorithms}
Table ~\ref{tab:fedvg_integration} presents numeric results for FedVG's integration with different federated learning methods. The experiments cover three datasets: CIFAR-10 and OrganAMNIST (using ResNet-18), and COVID-19 (using ResNet-50), which are evaluated across a range of data heterogeneity settings. Each baseline FL algorithm, including FedAvg, FedAvgM, FedProx, Scaffold, FedDyn, and Elastic, is assessed both alone and combined with FedVG.

\begin{table*}[t]
\centering
\caption{Performance comparison of federated learning methods and their integration with FedVG on CIFAR-10, OrganAMNIST, and COVID-19 datasets using ResNet-18 and ResNet-50 models across data heterogeneity levels ($\alpha$). \textbf{Bold} values indicate best accuracies and \underline{underlined} values indicate second best accuracies.}
\begin{tabular}{l|ccccc}
\toprule
Method & \textbf{$\alpha=100$} & \textbf{$\alpha=10$} & \textbf{$\alpha=1$} & \textbf{$\alpha=0.1$} & \textbf{$\alpha=0.05$} \\
\midrule
\multicolumn{6}{c}{\textit{CIFAR-10 / ResNet-18}} \\
\midrule
FedAvg & 80.37 ± 0.89 & 77.95 ± 0.43 & \textbf{76.26 ± 1.65} & 57.93 ± 1.82 & 48.83 ± 5.53 \\
FedVG & \underline{80.74 ± 0.79} & \textbf{79.65 ± 0.52} & \underline{75.69 ± 2.27} & \textbf{61.06 ± 0.34} & \textbf{53.58 ± 2.78} \\
FedAvg + FedVG & \textbf{80.92 ± 0.79} & \underline{78.79 ± 0.56} & 74.96 ± 2.55 & \underline{60.60 ± 2.26} & \underline{52.49 ± 3.83} \\
\midrule
FedAvgM & 71.09 ± 5.16 & 56.64 ± 10.75 & \textbf{51.87 ± 11.40} & 34.18 ± 4.38 & 25.87 ± 7.04 \\
FedAvgM + FedVG & \textbf{77.87 ± 1.98} & \textbf{63.05 ± 12.55} & 39.51 ± 17.02 & \textbf{42.63 ± 7.25} & \textbf{33.25 ± 3.62} \\
\midrule
FedProx & 79.92 ± 1.50 & 77.39 ± 1.32 & \textbf{76.73 ± 0.63} & 55.22 ± 5.56 & 49.19 ± 6.55 \\
FedProx + FedVG & \textbf{81.16 ± 0.21} & \textbf{79.10 ± 1.85} & 74.60 ± 1.75 & \textbf{60.18 ± 2.18} & \textbf{52.67 ± 3.12} \\
\midrule
Scaffold & \textbf{81.05 ± 0.16} & 80.39 ± 0.15 & \textbf{79.25 ± 0.23} & 56.15 ± 2.65 & \textbf{41.94 ± 1.91} \\
Scaffold + FedVG & 80.87 ± 0.15 & \textbf{80.42 ± 0.15} & 79.22 ± 0.18 & \textbf{56.59 ± 1.28} & 39.94 ± 3.13 \\
\midrule
FedDyn & 66.37 ± 0.33 & 65.90 ± 0.36 & 62.37 ± 0.29 & \textbf{41.66 ± 1.22} & \textbf{39.26 ± 1.99} \\
FedDyn + FedVG & \textbf{68.62 ± 0.32} & \textbf{68.27 ± 0.39} & \textbf{63.34 ± 0.52} & 37.25 ± 0.32 & 32.59 ± 0.74 \\
\midrule
Elastic & 79.45 ± 0.27 & 76.55 ± 1.88 & 75.69 ± 0.89 & 55.86 ± 4.26 & \textbf{54.92 ± 2.16} \\
Elastic + FedVG & \textbf{79.95 ± 0.23} & \textbf{79.34 ± 0.62} & \textbf{76.46 ± 0.97} & \textbf{59.09 ± 4.07} & 51.75 ± 4.49 \\
\midrule
\multicolumn{6}{c}{\textit{OrganAMNIST / ResNet-18}} \\
\midrule
FedAvg & 98.46 ± 0.16 & 98.87 ± 0.21 & 98.47 ± 0.24 & 93.50 ± 1.22 & 86.37 ± 2.49 \\
FedVG & \textbf{99.41 ± 0.08} & \textbf{99.42 ± 0.02} & \textbf{99.12 ± 0.10} & \textbf{94.72 ± 0.68} & \textbf{87.57 ± 1.91} \\
FedAvg + FedVG & \underline{98.83 ± 0.29} & \underline{99.22 ± 0.11} & \underline{98.76 ± 0.30} & \underline{94.35 ± 1.24} & \underline{87.09 ± 1.25} \\
\midrule
FedAvgM & 87.14 ± 4.56 & 93.43 ± 4.13 & 94.00 ± 4.76 & 78.95 ± 8.14 & 60.15 ± 7.61 \\
FedAvgM + FedVG & \textbf{99.20 ± 0.07} & \textbf{99.52 ± 0.06} & \textbf{98.77 ± 0.70} & \textbf{88.87 ± 5.06} & \textbf{77.01 ± 7.10} \\
\midrule
FedProx & 98.44 ± 0.21 & 98.68 ± 0.32 & 98.60 ± 0.20 & 93.19 ± 1.63 & 83.80 ± 1.25 \\
FedProx + FedVG & \textbf{99.44 ± 0.05} & \textbf{99.43 ± 0.05} & \textbf{99.02 ± 0.17} & \textbf{94.97 ± 0.37} & \textbf{88.00 ± 1.24} \\
\midrule
Scaffold & \textbf{99.22 ± 0.06} & \textbf{99.17 ± 0.06} & \textbf{98.85 ± 0.06} & 93.11 ± 0.62 & \textbf{84.32 ± 1.72} \\
Scaffold + FedVG & 99.14 ± 0.10 & \textbf{99.17 ± 0.07} & 98.80 ± 0.06 & \textbf{93.19 ± 0.27} & 83.19 ± 4.53 \\
\midrule
FedDyn & 97.26 ± 0.17 & 97.13 ± 0.07 & 95.60 ± 0.15 & \textbf{78.61 ± 1.01} & \textbf{68.17 ± 0.62} \\
FedDyn + FedVG & \textbf{97.79 ± 0.08} & \textbf{97.78 ± 0.07} & \textbf{96.24 ± 0.23} & 77.25 ± 1.58 & 63.31 ± 1.88 \\
\midrule
Elastic & 99.12 ± 0.11 & 98.78 ± 0.18 & 97.94 ± 0.27 & 88.46 ± 3.70 & 79.96 ± 5.20 \\
Elastic + FedVG & \textbf{99.31 ± 0.05} & \textbf{99.33 ± 0.08} & \textbf{99.00 ± 0.08} & \textbf{92.98 ± 1.39} & \textbf{87.95 ± 1.00} \\
\midrule
\multicolumn{6}{c}{\textit{COVID-19 / ResNet-50}} \\
\midrule
FedAvg & 88.23 ± 0.23 & \underline{87.88 ± 0.38} & \textbf{87.91 ± 0.37} & \textbf{84.34 ± 0.67} & 64.92 ± 5.45 \\
FedVG & \textbf{88.30 ± 0.66} & 87.83 ± 0.39 & 87.40 ± 0.74 & 83.10 ± 0.54 & \textbf{75.18 ± 1.36} \\
FedAvg + FedVG & \underline{88.26 ± 0.53} & \textbf{88.38 ± 0.54} & \underline{87.88 ± 0.59} & \underline{83.62 ± 0.83} & \underline{68.39 ± 4.12} \\
\midrule
FedAvgM & \textbf{90.16 ± 0.23} & \textbf{90.23 ± 0.54} & 90.00 ± 0.33 & 85.54 ± 0.87 & 60.07 ± 9.31 \\
FedAvgM + FedVG & 89.96 ± 0.46 & 89.88 ± 0.55 & \textbf{90.02 ± 0.25} & \textbf{88.03 ± 0.52} & \textbf{72.66 ± 4.18} \\
\midrule
FedProx & 87.93 ± 0.55 & 87.91 ± 0.37 & \textbf{87.53 ± 0.47} & \textbf{84.06 ± 1.06} & 64.58 ± 5.34 \\
FedProx + FedVG & \textbf{88.22 ± 0.22} & \textbf{88.39 ± 0.46} & 87.32 ± 0.53 & 83.43 ± 0.80 & \textbf{72.55 ± 2.54} \\
\midrule
Scaffold & \textbf{88.70 ± 0.30} & 88.67 ± 0.55 & 88.36 ± 0.26 & \textbf{84.47 ± 1.06} & 70.05 ± 5.29 \\
Scaffold + FedVG & 88.63 ± 0.28 & \textbf{88.88 ± 0.40} & \textbf{88.64 ± 0.28} & 83.77 ± 0.93 & \textbf{71.62 ± 3.74} \\
\midrule
FedDyn & 84.48 ± 0.48 & \textbf{84.51 ± 0.33} & 84.28 ± 0.75 & \textbf{78.49 ± 0.48} & 63.01 ± 5.79 \\
FedDyn + FedVG & \textbf{88.30 ± 0.66} & 84.03 ± 0.44 & \textbf{84.31 ± 0.74} & 76.32 ± 3.57 & \textbf{66.99 ± 3.98} \\
\midrule
Elastic & \textbf{87.50 ± 0.37} & \textbf{87.12 ± 0.57} & 86.69 ± 0.78 & \textbf{83.09 ± 0.79} & 65.87 ± 3.69 \\
Elastic + FedVG & 86.99 ± 0.44 & 86.71 ± 0.44 & \textbf{86.95 ± 0.75} & 82.70 ± 0.68 & \textbf{75.24 ± 2.12} \\
\bottomrule
\end{tabular}
\label{tab:fedvg_integration}
\end{table*}

\section{Details of Ablation Analysis}

\subsection{Analysis of Global Validation Set $D_{val}$}
\mbox{}\\
\indent


\begin{figure*}
     \centering
     \begin{subfigure}[t]{0.24\textwidth}
         \centering
         \includegraphics[width=\textwidth]{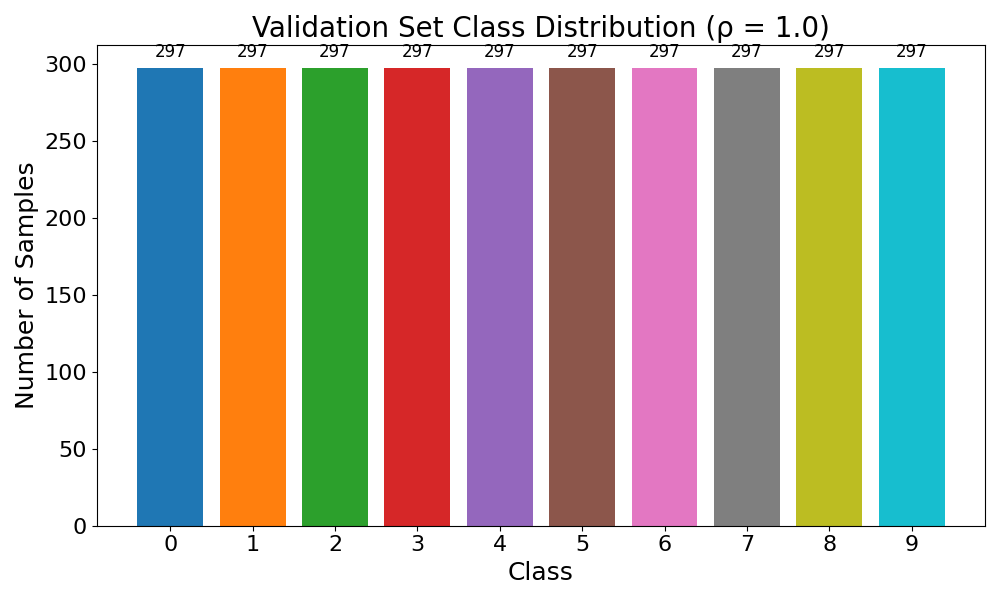}
         \caption{$\rho = 1$}
         \label{fig:rho_1}
     \end{subfigure}
     \hfill
     \begin{subfigure}[t]{0.24\textwidth}
         \centering
         \includegraphics[width=\textwidth]{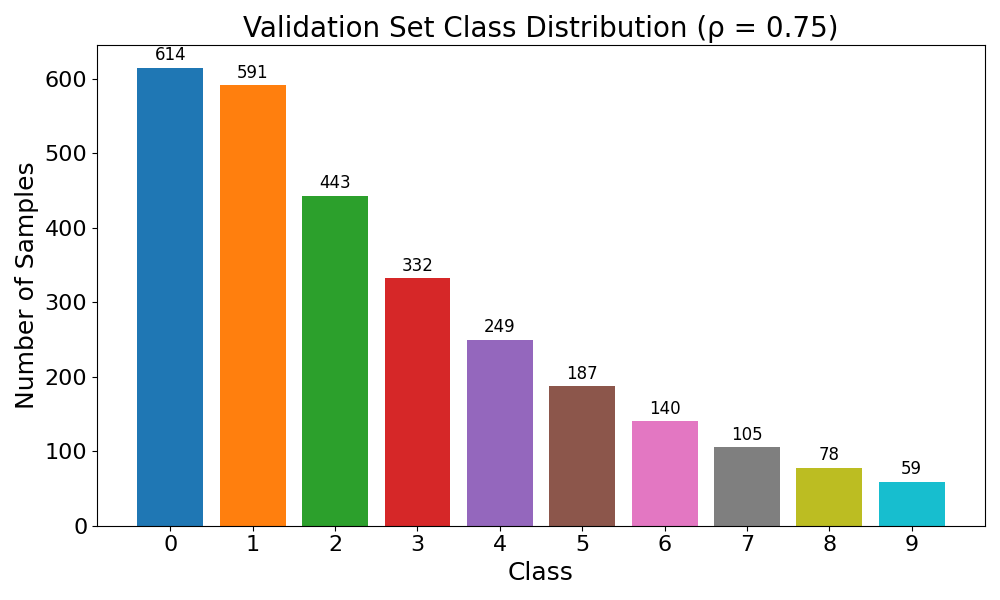}
         \caption{$\rho = 0.75$}
         \label{fig:rho_0.75}
     \end{subfigure}
     \hfill
     \begin{subfigure}[t]{0.24\textwidth}
         \centering
         \includegraphics[width=\textwidth]{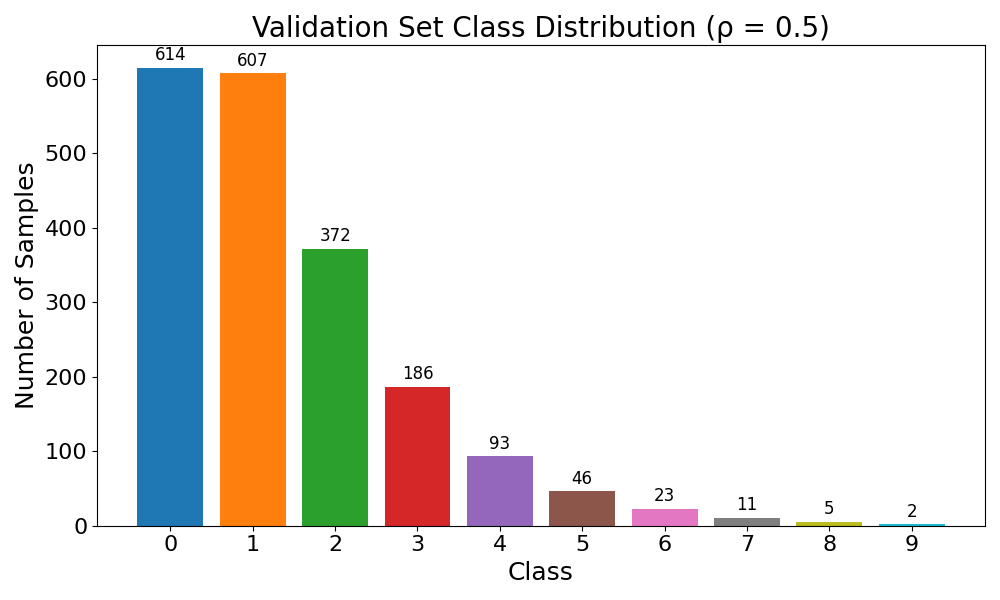}
         \caption{$\rho = 0.5$}
         \label{fig:rho_0.5}
     \end{subfigure}
     \hfill
     \begin{subfigure}[t]{0.24\textwidth}
         \centering
         \includegraphics[width=\textwidth]{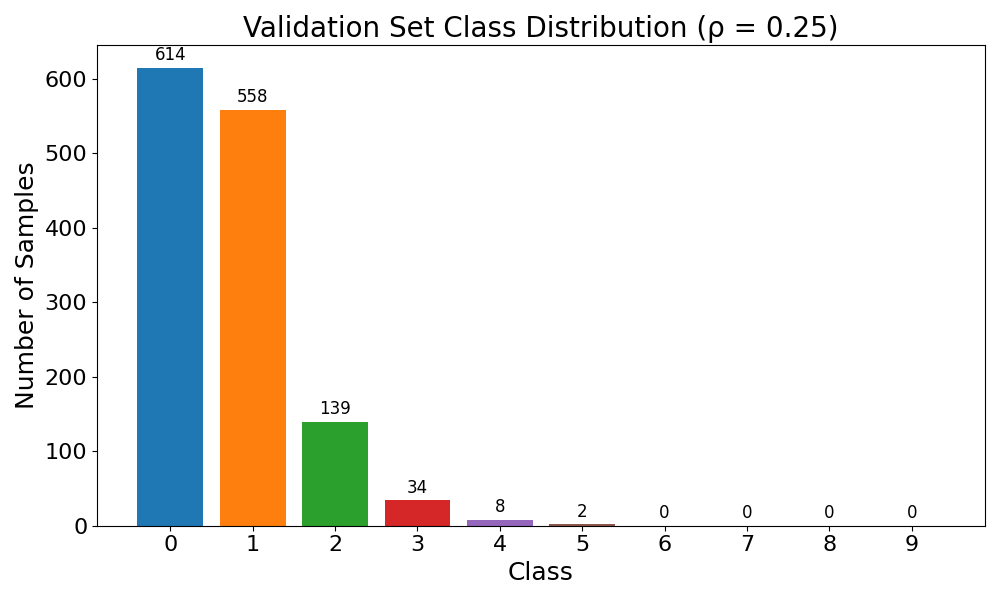}
         \caption{$\rho = 0.25$}
         \label{fig:rho_0.25}
     \end{subfigure}
        \caption{Visualization of class imbalance of $D_{val}$ at various values of $\rho$.}
        \label{fig:class_imbalance_examples}
\end{figure*}

To simulate varying levels of class imbalance in the global validation set, we employ a controlled sampling strategy based on the imbalance ratio, $\rho \in \left(0, 1\right]$, where $\rho = 1$ indicates a balanced class distribution.  To ensure a fair comparison, we fix the total size of the imbalanced validation set to half the size of the original validation set, i.e., $N' = \frac{1}{2}\left| D_{val} \right| $.

First we compute the unnormalized class proportion as:
\begin{equation}
    p_i' = \rho^i, \quad i = 0, 1, \ldots, C-1.
\end{equation}

These are then normalized to obtain valid class probabilities so that $\sum_{i=0}^{C-1}p_i =1 $.
\begin{equation}
    p_i = \frac{\rho^{i}}{\sum_{j=0}^{C-1} \rho^{j}}
\end{equation}

Finally, the number of validation samples allocated to each class is given by $s_i = \left\lfloor p_i \times N' \right\rfloor$, where, $\lfloor . \rfloor$ denotes the floor operation.

Figure ~\ref{fig:class_imbalance_examples} shows examples of class distributions for $D_{val}$ of the CIFAR-10 dataset, as the imbalance ratio $\rho \xrightarrow{} 0$.




\subsection{Evaluation of norm type}

\begin{figure}[t]
\centering
\includegraphics[width=1\columnwidth]{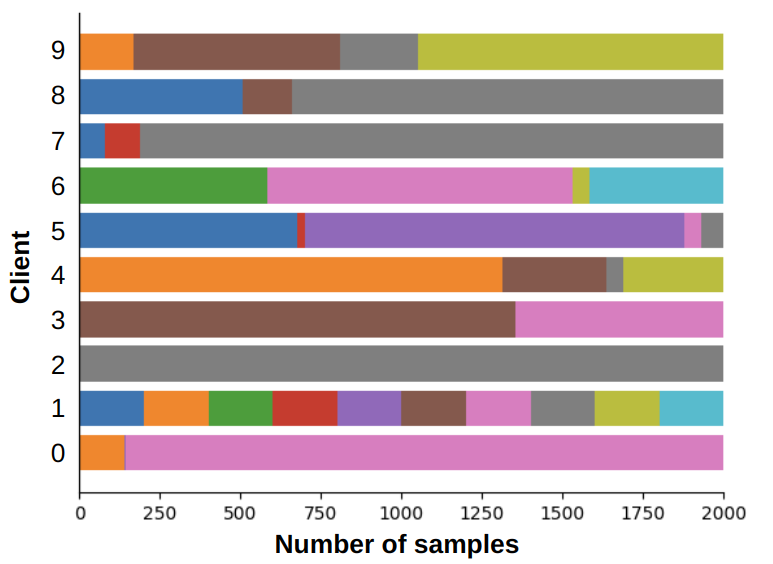}
\caption{Visualization of sample distributions used in the norm-type ablation study. Here, client one is configured to be perfectly balanced.}
\label{fig:1balanced}
\end{figure}

Eqn. ~\ref{eq:general_norm} represents a general norm for converting layer-wise gradients into a mean gradient value, where $\left\| \cdot \right\|$ denotes a general vector or matrix norm. Common choices for $\left\| \cdot \right\|$ include $\ell_1$-norm, $\ell_2$-norm, and spectral norm (i.e., the largest singular value of the gradient matrix), etc.

Formally, for a vector $\mathbf{g} = (g_1, g_2, \ldots, g_d) \in \mathbb{R}^d$,

\begin{itemize}
    \item The $\ell_1$-norm is defined as $\|\mathbf{g}\|_1 = \sum_{i=1}^d |g_i|$.
    \item The $\ell_2$-norm is defined as $\|\mathbf{g}\|_2 = \sqrt{\sum_{i=1}^d g_i^2}$.
\end{itemize}

For a matrix $G \in \mathbb{R}^{m \times n}$, representing the gradient of a layer,
\begin{itemize}
    \item the spectral norm $\|G\|_{\sigma}$ is defined as the largest singular value of $G$, which corresponds to $\|G\|_{\sigma} = \sigma_{\max}(G)$
    \item where, \(\sigma_{\max}(G)\) is the maximum singular value obtained from the singular value decomposition (SVD) of$G$.
\end{itemize}


When calculating the $\ell_1$-norm and $\ell_2$-norm, matrices are first flattened into vectors. For the spectral norm, only layer gradients with two or more dimensions are considered and reshaped into matrices for singular value decomposition.

In addition, we define the \textit{delta norm}, which is computed as the norm of parameter updates (i.e., the difference between client models and the global model) rather than the layer gradients. For \textit{delta norm}, Eqn.~\ref{eq:general_norm} becomes

\begin{equation}
    \bar{G}_k = \frac{1}{L}\sum_{\ell=1}^{L}{\left\| \theta_g^{l} - \theta_k^{l} \right\|}.
\end{equation}

Figure ~\ref{fig:1balanced} shows the client setup for these experiments. Here, 9 clients contain the same number of elements in their local dataset, with a concentration coefficient $\alpha = 0.05$. The tenth client (client 1) is made to be perfectly homogeneous. Figure ~\ref{fig:norm_abl} then shows the FedVG weights assigned to each client at each round of federated training.

\subsection{Aggregation granularity}
To analyze the impact of aggregation granularity, we evaluate three strategies: modelwise, layerwise, and blockwise aggregation.

In the modelwise setting (the default FedVG configuration), the client’s mean gradient magnitude is computed by averaging gradient norms across all $L$ layers

\begin{equation}
    \bar{G}_k = \frac{1}{L}\sum_{\ell=1}^{L}{\left\|  \nabla_{\theta_k^{(\ell)}} \mathcal{L}_{\mathrm{val}} \right\|}.
\end{equation}

and the corresponding raw score is

\begin{equation}
    \hat{s_k} = \frac{1}{\bar{G_k} + \epsilon}
\end{equation}

In the layerwise variant, we compute the raw score ($\hat{s_k}^{(l)}$) and normalized score ($s_k^{(l)}$) for each layer $l$ separately. The raw and normalized scores are given by:
\begin{align}
    G_k^{(l)} = \left\|  \nabla_{\theta_k^{(\ell)}} \mathcal{L}_{\mathrm{val}} \right\|
    \\
    \hat{s_k}^{(l)} = \frac{1}{G_k^{(l)} + \epsilon}
    \\
    s_k^{(l)} = \frac{\hat{s_k}^{(l)}}{\sum_{j=1}^{K} \hat{s_j}^{(l)}}
\end{align}
The aggregation step is then performed for each layer:
\begin{equation}
    \theta_g^{(l)} \leftarrow \theta_g^{(l)} - \sum_{k=1}^{K} s_k^{(l)} (\theta_g^{(l)} - \theta_k^{(l)}) ,
    \quad \forall\, l = 1, 2, \dots, L
\end{equation}

In the blockwise variant, we follow a similar approach to modelwise aggregation, except the averaging is performed over layers within each block rather than the entire model. For ResNet models, each residual block is treated as a block; for ViT-based models, each transformer encoder block is treated as a block; and in both cases, the classifier head is considered a separate block. The model $\theta_k$ can be decomposed into $B$ blocks,
\begin{equation}
    \theta_k = \{\theta_k^{(1)}, \theta_k^{(2)}, \ldots, \theta_k^{(B)}\}
\end{equation}
Let $|b|$ denote the number of layers in block $b$.  The block-level score computation is:
\begin{align}
    \bar{G}_k^{(b)} = \frac{1}{|b|}\sum_{\ell \in b}{\left\|  \nabla_{\theta_k^{(\ell)}} \mathcal{L}_{\mathrm{val}} \right\|}
    \\
    \hat{s_k}^{(b)} = \frac{1}{G_k^{(b)} + \epsilon}
    \\
    s_k^{(b)} = \frac{\hat{s_k}^{(b)}}{\sum_{j=1}^{K} \hat{s_j}^{(b)}}
\end{align}
The aggregation is then applied at the block level:
\begin{equation}
        \theta_g^{(b)} \leftarrow \theta_g^{(b)} - \sum_{k=1}^{K} s_k^{(b)} (\theta_g^{(b)} - \theta_k^{(b)}) ,
    \quad \forall\, b = 1, 2, \dots, B.
\end{equation}

Table ~\ref{tab:aggregation_granularity} reports the detailed performance of the three FedVG variants (Modelwise, Layerwise, and Blockwise), across different dataset–model combinations and heterogeneity levels ($\alpha \in {100, 10, 1, 0.1, 0.05}$).

\begin{table*}[t]
\centering
\caption{
Comparison of FedVG variants (Modelwise, Layerwise, and Blockwise) across datasets, models, and heterogeneity levels ($\alpha$ values). \textbf{Bold} values indicate best accuracies.
}
\begin{tabular}{l|ccccc}
\toprule
Method & \textbf{$\alpha=100$} & \textbf{$\alpha=10$} & \textbf{$\alpha=1$} & \textbf{$\alpha=0.1$} & \textbf{$\alpha=0.05$} \\
\midrule

\multicolumn{6}{c}{\textit{CIFAR-10 / ResNet-18}} \\
\midrule
FedVG (Modelwise) & \textbf{80.74 ± 0.79} & \textbf{79.65 ± 0.52} & \textbf{75.69 ± 2.27} & \textbf{61.06 ± 0.34} & \textbf{53.58 ± 2.78} \\
FedVG (Layerwise) & 77.87 ± 2.79 & 73.33 ± 5.30 & 74.03 ± 2.41 & 59.43 ± 2.24 & 50.25 ± 1.80 \\
FedVG (Blockwise) & 79.33 ± 1.85 & 71.15 ± 8.79 & 71.85 ± 2.96 & 58.06 ± 3.87 & 50.30 ± 2.95 \\
\midrule

\multicolumn{6}{c}{\textit{OrganAMNIST / ResNet-18}} \\
\midrule
FedVG (Modelwise) & \textbf{99.41 ± 0.08} & \textbf{99.42 ± 0.02} & \textbf{99.12 ± 0.10} & \textbf{94.72 ± 0.6} & \textbf{87.57 ± 1.91} \\
FedVG (Layerwise) & 99.33 ± 0.09 & 99.30 ± 0.29 & 98.94 ± 0.17 & 94.30 ± 1.86 & 87.44 ± 0.99 \\
FedVG (Blockwise) & 99.29 ± 0.06 & 99.39 ± 0.06 & 99.02 ± 0.16 & 92.83 ± 2.16 & 87.02 ± 1.42 \\
\midrule

\multicolumn{6}{c}{\textit{Tiny ImageNet / ResNet-50}} \\
\midrule
FedVG (Modelwise) & 69.21 ± 0.10 & 69.09 ± 0.05 & 68.93 ± 0.18 & 67.21 ± 0.27 & \textbf{65.42 ± 0.18} \\
FedVG (Layerwise) & 69.23 ± 0.15 & \textbf{69.12 ± 0.09} & 68.79 ± 0.45 & 66.85 ± 0.56 & 64.88 ± 0.99 \\
FedVG (Blockwise) & \textbf{69.31 ± 0.15} & \textbf{69.12 ± 0.11} & \textbf{69.02 ± 0.13} & \textbf{67.27 ± 0.19} & 65.09 ± 0.79 \\
\midrule

\multicolumn{6}{c}{\textit{COVID-19 / ResNet-50}} \\
\midrule
FedVG (Modelwise) & 88.30 ± 0.66 & 87.83 ± 0.39 & 87.40 ± 0.74 & 83.10 ± 0.54 & \textbf{75.18 ± 1.36} \\
FedVG (Layerwise) & \textbf{87.88 ± 0.51} & \textbf{87.88 ± 0.36} & 87.42 ± 0.54 & 83.71 ± 0.78 & 73.44 ± 1.74 \\
FedVG (Blockwise) & 88.38 ± 0.59 & 88.23 ± 0.42 & \textbf{87.75 ± 0.54} & \textbf{83.54 ± 0.90} & 74.23 ± 1.73 \\
\midrule

\multicolumn{6}{c}{\textit{DermaMNIST / ResNet-50}} \\
\midrule
FedVG (Modelwise) & \textbf{83.66 ± 0.31} & \textbf{83.45 ± 0.31} & 83.58 ± 0.55 & 77.82 ± 0.46 & 71.31 ± 0.71 \\
FedVG (Layerwise) & 83.46 ± 0.27 & 83.08 ± 0.40 & 83.19 ± 0.51 & \textbf{77.83 ± 0.48} & 71.22 ± 0.72 \\
FedVG (Blockwise) & 83.53 ± 0.65 & 83.21 ± 0.17 & \textbf{83.79 ± 0.18} & 77.78 ± 0.54 & \textbf{71.48 ± 1.00} \\
\midrule

\multicolumn{6}{c}{\textit{COVID-19 / ViT-S/16}} \\
\midrule
FedVG (Modelwise) & 88.29 ± 0.33 & \textbf{89.06 ± 0.29} & 89.26 ± 0.45 & \textbf{87.47 ± 0.78} & 83.34 ± 1.53 \\
FedVG (Layerwise) & \textbf{88.41 ± 0.68} & 88.66 ± 0.73 & 89.57 ± 0.32 & 87.17 ± 0.58 & 83.45 ± 1.92 \\
FedVG (Blockwise) & 87.98 ± 0.69 & 88.98 ± 0.20 & \textbf{89.66 ± 0.63} & 87.01 ± 0.61 & \textbf{83.90 ± 1.74} \\
\midrule

\multicolumn{6}{c}{\textit{DermaMNIST / ViT-B/16}} \\
\midrule
FedVG (Modelwise) & 80.46 ± 2.31 & \textbf{81.60 ± 0.26} & 81.46 ± 0.50 & 78.31 ± 0.99 & 76.20 ± 0.76 \\
FedVG (Layerwise) & \textbf{80.72 ± 1.60} & 81.35 ± 0.43 & \textbf{81.47 ± 0.26} & \textbf{79.25 ± 0.80} & 73.92 ± 2.51 \\
FedVG (Blockwise) & 79.70 ± 2.87 & 81.20 ± 0.56 & 79.63 ± 2.89 & 79.05 ± 0.85 & \textbf{76.29 ± 0.67} \\
\bottomrule
\end{tabular}
\label{tab:aggregation_granularity}
\end{table*}